\RequirePackage{fix-cm}
\documentclass[twocolumn]{svjour3}          
\smartqed  
\usepackage{amsmath}
\usepackage{gensymb}
\usepackage{diagbox}
\usepackage{threeparttable}
\usepackage{xspace}
\usepackage{siunitx}
\usepackage{graphicx}
\usepackage{subfig}
\usepackage{placeins}
\usepackage{blindtext}
\usepackage{amssymb}
\usepackage{pifont}
\usepackage{threeparttable}
\usepackage{color}
\usepackage[normalem]{ulem}
\usepackage{multirow}
\usepackage{float}
\usepackage{amsfonts}
\usepackage{bm}
\usepackage{bbm}
\usepackage{array}
\usepackage{tabulary}	
\usepackage{microtype}
\usepackage{booktabs}
\usepackage{xspace}
\usepackage[table]{xcolor}
\usepackage{colortbl}
\usepackage{diagbox}
\usepackage{rotating}
\usepackage{booktabs}
\usepackage{overpic}
\usepackage{enumitem}
\usepackage{colortbl}
\usepackage{soul}
\usepackage{url}
\usepackage{makecell, verbatim, sidecap}
\usepackage{algorithm}
\usepackage{cite}
\usepackage[T1]{fontenc}
\usepackage{bbm}
\usepackage{tikz}
\usepackage{comment}
\usepackage{graphicx}
\usepackage{amssymb}
\usepackage{booktabs}
\usepackage{algpseudocode}
\usepackage[accsupp]{axessibility} 
\usepackage{xfrac}
\usepackage{color}
\usepackage{wrapfig}
\usepackage{microtype}
\usepackage{tabularx}
\usepackage{booktabs}
\usepackage{multirow}
\usepackage{blindtext}
\usepackage{xspace}
\usepackage[pagebackref=true,breaklinks=true,letterpaper=true,colorlinks,bookmarks=false]{hyperref}

\makeatletter
\newcommand{\thickhline}{%
    \noalign {\ifnum 0=`}\fi \hrule height 1pt
    \futurelet \reserved@a \@xhline
}
\makeatother

\makeatletter
\DeclareRobustCommand\onedot{\futurelet\@let@token\@onedot}
\def\@onedot{\ifx\@let@token.\else.\null\fi\xspace}

\makeatother

\newcommand{\app}{\raise.17ex\hbox{$\scriptstyle\sim$}}

\journalname{IJCV}
\begin{document}
\title{RegFormer++: An Efficient Large-Scale 3D LiDAR Point Registration Network with Projection-Aware 2D Transformer}
\sloppy


\author{Jiuming Liu$^{\dagger}$ $^{1}$ \and
        Guangming Wang$^{\dagger}$ $^{2}$ \and
        Zhe Liu $^{1}$ \and
        Chaokang Jiang $^3$ \and\\
        Haoang Li $^4$ \and 
        Mengmeng Liu $^5$ \and
        Tianchen Deng $^1$ \and
        Marc Pollefeys $^6$ \and \\
        Michael Ying Yang $^7$ \and
        Hesheng Wang$^{*}$ $^1$
}

\institute{$^{*}$Corresponding author: Hesheng Wang. (email: wanghesheng@sjtu.edu.cn)
\\
$^{\dagger}$The first two authors contributed equally.\\
This work was supported in part by National Key R\&D Program of China under Grant No.2024YFB4708900, and also in part by the Natural Science Foundation of China under Grant 62225309, U24A20278, 62361166632, and U21A20480.
$^{1}$Jiuming Liu, Zhe Liu, Tianchen Deng, and Hesheng Wang are with the School of Automation and Intelligent Sensing, State Key Laboratory of Avionics Integration and Aviation System-of-Systems Synthesis, Shanghai Jiao Tong University, Shanghai 200240, China. (email:liujiuming123@gmail.com) \\
$^{2}$Guangming Wang is with the Department of Engineering, University of Cambridge, UK. \\ 
$^{3}$Chaokang Jiang is with China University of Mining and Technology. \\
$^{4}$Haoang Li is with the Thrust of Robotics and Autonomous Systems and the Thrust of Intelligent Transportation, Hong Kong University of Science and Technology (Guangzhou), Guangzhou 511453, China. \\ 
$^{5}$Mengmeng Liu is with University of Twente.\\
$^{6}$Marc Pollefeys is with both ETH Zurich and Microsoft.\\
$^{7}$Michael Ying Yang is with University of Bath.    
}

\date{Received: xx Mar 2025 / Accepted: xx xx 2025}

\maketitle

\begin{abstract}
 Although point cloud registration has achieved remarkable advances in object-level and indoor scenes, large-scale LiDAR registration methods has been rarely explored before. Challenges mainly arise from the huge point scale, complex point distribution, and numerous outliers within outdoor LiDAR scans. In addition, most existing registration works generally adopt a two-stage paradigm: They first find correspondences by extracting discriminative local descriptors and then leverage robust estimators (e.g. RANSAC) to filter outliers, which are highly dependent on well-designed descriptors and post-processing choices. To address these problems, we propose a novel end-to-end differential transformer network, termed RegFormer++, for large-scale point cloud alignment without requiring any further post-processing. Specifically, a hierarchical projection-aware 2D transformer with linear complexity is proposed to project raw LiDAR points onto a cylindrical surface and extract global point features, which can improve resilience to outliers due to long-range dependencies. Because we fill original 3D coordinates into 2D projected positions, our designed transformer can benefit from both high efficiency in 2D processing and accuracy from 3D geometric information. Furthermore, to effectively reduce wrong point matching, a Bijective Association Transformer (BAT) is designed, combining both cross attention and all-to-all point gathering. To improve training stability and robustness, a feature-transformed optimal transport module is also designed for regressing the final pose transformation. Extensive experiments on KITTI, NuScenes, and Argoverse datasets demonstrate that our model achieves state-of-the-art performance in terms of both accuracy and efficiency. Also, our RegFormer++ can generalize well on the unseen LiDAR data with diverse sensor settings, e.g. ETH dataset. Additionally, we apply our registration network into the real-time LiDAR odometry task, with competitive 0.48\degree/100m  average rotational RMSE and 0.96\% translational errors. Codes are available at \url{https://github.com/IRMVLab/RegFormer}.
\end{abstract}

\section{Introduction}\label{sec:introduction}

\begin{figure*}
  \centering
  \includegraphics[width=0.9\linewidth]{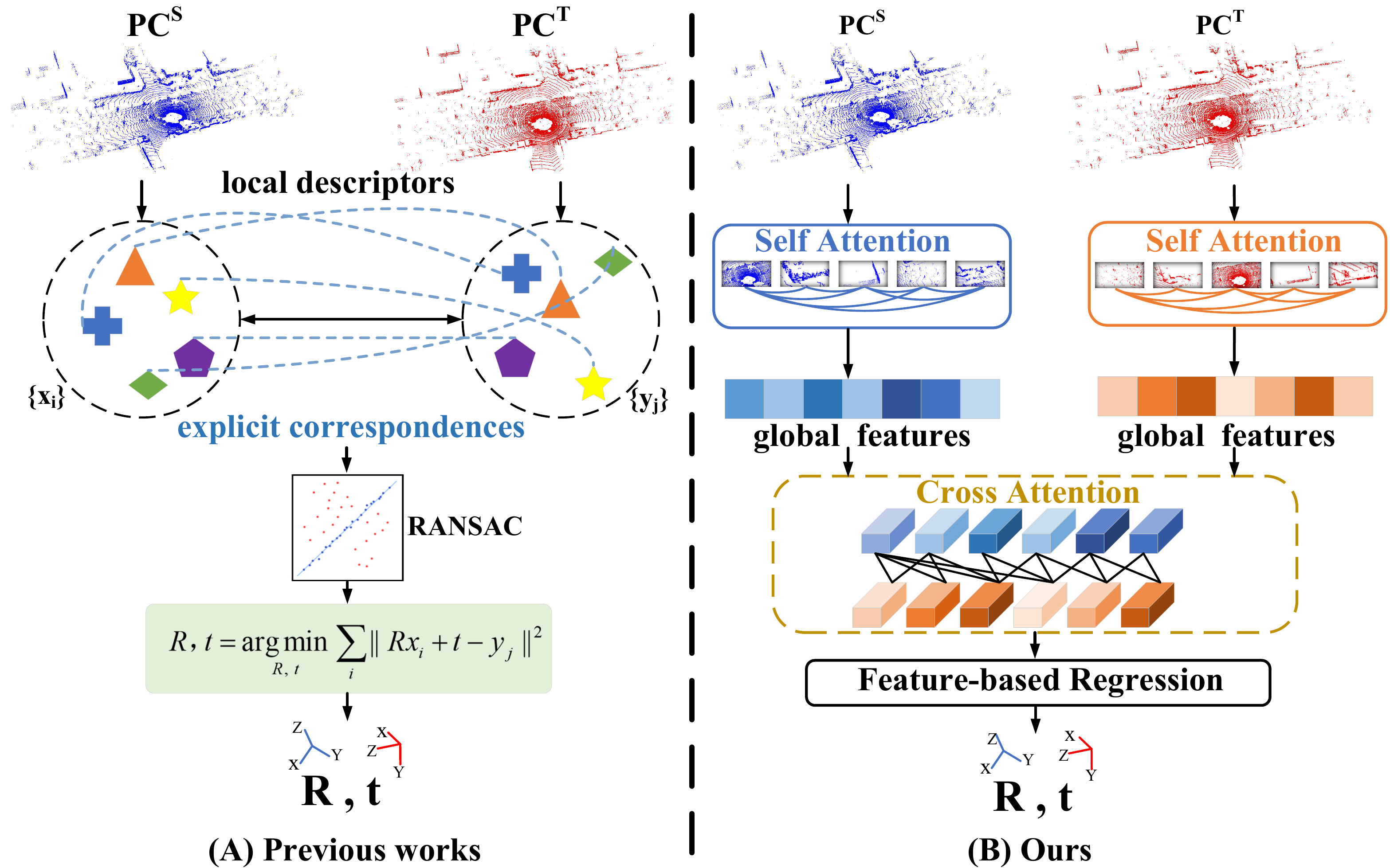}
  \vspace{-3mm}
  \caption{Comparison with previous point cloud registration works. Previous methods (A) extract local descriptors and establish explicit correspondences, while our RegFormer++ rely on globally-aware features and learn cross-frame association by cross attention requiring no correspondences. Furthermore, prior literature resorts to the RANSAC \cite{fischler1981random} algorithm for outlier filtering. In contrast, our method leverage attention mechanism to softly remove outliers. }
  \label{fig:compipe}
  
\end{figure*}

Point cloud registration is a fundamental problem in 3D computer vision, which aims to estimate the rigid transformation between two point cloud frames. It has been widely applied in autonomous driving \cite{li2024intelligent,zeng2025FSDrive,zeng2025janusvln}, SLAM systems \cite{liu2024dvlo,deng2024compact,zhu2024sni}, and medical images \cite{chen2021data,du2024embracing}.

Although learning-based methods show great potential in object-level or indoor registration tasks \cite{yew20183dfeat,choy2019fully,ao2021spinnet,huang2021predator}, large-scale point cloud registration is still less studied. Challenges are mainly three-fold: 1) Outdoor LiDAR scans consist of numerous unstructured points, which are intrinsically sparse, irregular, and have a large spatial range. It is non-trivial to efficiently process all raw points in one inference \cite{yew2022regtr,liu2025topolidm}. 2) Outliers from dynamic objects and occlusion would degrade the registration accuracy as they introduce uncertain and inconsistent motions \cite{wang2021pwclo,ma2025ms}. 3) There are mismatches (wrong point matching) when directly leveraging distance-based nearest neighbor matching methods (eg. $k$NN) to distant point cloud pairs \cite{lu2021hregnet}, especially for low-overlap inputs.

\textbf{For the first challenge}, previous registration works mostly voxelize input points \cite{bai2020d3feat,huang2021predator,yew2022regtr,tao2025lsreg}, and then establish putative correspondences by selecting keypoints and learning distinctive local descriptors \cite{yew20183dfeat,choy2019fully,ao2021spinnet,ma2026gor} as in Fig.~\ref{fig:compipe}. However, quantization errors are inevitable in the voxelization process \cite{huang20223qnet,liu2024point}. Keypoint selection strategy may also influence registration accuracy and the downsampling process challenges repeatability \cite{yu2021cofinet}. In this paper, instead of searching keypoints, we directly process all dense LiDAR points by projecting them onto a cylindrical surface for the structured organization. The projected image-like structure also facilitates the subsequent window-based 2D attention, which enables our network to efficiently process up to 120000 points with linear computational costs. To take advantage of 3D geometric features, each projected 2D position is filled with raw 3D point coordinates. Another concern is that projected pseudo images are full of invalid positions due to the original sparsity of point clouds. We handle this by designing a projection-aware mask, which represents whether each projected position has a corresponding point in raw point clouds.  

\textbf{For the second challenge}, the commonly used method is applying RANSAC \cite{fischler1981random,bai2020d3feat,ao2021spinnet,huang2021predator} to filter outliers as in Fig.~\ref{fig:compipe}. However, RANSAC suffers from slow convergence \cite{qin2022geometric} and also highly relies on post-processing choices \cite{yew2022regtr}. In this paper, we rethink the outlier issue and observe that global modeling capability is critical to localize occluded objects and recognize dynamics as they introduce inconsistent global motion \cite{yu2024riga,liu2025mamba4d,liu2025dvlo4d,liu20254dstr}. Therefore, we propose a projection-aware transformer to extract point features globally. Notably, some recent works \cite{qin2022geometric,yew2022regtr,fu2025dual} also try to design RANSAC-free registration networks. However, their feature extraction modules are still established on CNNs, without the global modeling capability. The closest approach to ours is REGTR \cite{yew2022regtr}, which directly predicts clean correspondences with the transformer. Nonetheless, the quadratic complexity of its transformer module deteriorates efficiency, which impairs large-scale application.

 \textbf{To tackle the third challenge}, a Bijective Association Transformer (BAT) is designed to reduce mismatches, combining a cross-attention layer and the all-to-all point gathering module. HRegNet \cite{lu2021hregnet,lu2023hregnet} already has awareness that nearest-neighbor matching can lead to considerable mismatches due to possible errors in descriptors. However, their $k$NN cluster is still distance-based, which can not deal well with low-overlap inputs. To address this problem, the cross-attention mechanism is utilized first for preliminary global location information exchange. Then, point features derived from the cross-attention module are delivered to the all-to-all point gathering module. Intuitively, features from deeper layers are coarse but reliable as they gather more information with larger receptive fields. Thus, we correlate each point in one frame with all points (rather than selecting $k$ points) in the other frame to gain reliable motion embeddings on the coarsest layer. Additionally, a feature-transformed optimal transport module is developed to strengthen the reliability of the matched point features. Finally, the precise transformation will be recovered by multiple iterative refinement layers.

This article is an extension of our previous work on large-scale point cloud registration published in \cite{Liu_2023_ICCV}. In summary, the contributions presented in \cite{Liu_2023_ICCV} are:
\begin{itemize}
	\item We propose an end-to-end pure transformer-based network for large-scale point cloud registration. The pipeline does not need any keypoint searching or post-processing, which is keypoint-free, correspondence-free, and RANSAC-free. 
	\item A projection-aware 2D transformer is designed to extract distinctive features with global modeling capability. Our transformer is also highly efficient, which can process up to 120000 points in real-time.
    \item A Bijective Association Transformer (BAT) module is developed to reduce mismatches by combining cross-attention with the all-to-all point correlation on the coarsest layer.
	
\end{itemize}

In this extended article, we propose following additional contributions:

\begin{itemize}
\item To further improve deterministic reliability and estimation stability, a feature-transformed optimal transport module is designed as in Section 3.6 and Algorithm 1. 
\item For enhanced efficiency, we introduced a projection-aware CUDA operator for parallel computing in \cite{Liu_2023_ICCV}. In this journal article, we extend more details about the design of this process in Section 3.3. 
\item To demonstrate the superiority of our method, additional quantitative on Argoverse dataset (Table 6) and qualitative experiments (Fig. 7, Fig. 8) are supplemented. Also, more implementation details of network settings and parameters are given in Table 1 and Table 2 for a better understanding. 
\item Furthermore, we evaluate the generalization ability of our method on unseen datasets with different sensor settings, i.e. ETH (Table 7), and also on a downstream task: LiDAR odometry (Table 8, Fig. 9, Fig. 10).
\item Our RegFormer++ further pushes the performance limits with 100\% successful registration recall on both KITTI \cite{geiger2012we,geiger2013vision} and NuScenes \cite{caesar2020nuscenes} datasets as in Table 3, Table 4, and Table 5. Compared to RegFormer, Regformer++ achieves nearly half of its estimation errors with 0.04m RTE and 0.14\degree RRE.
\end{itemize}

\section{Related Work}
Before delving into the network details, we first review recent advancements in four related topics: classical point cloud registration, learning-based point cloud registration, large-scale point cloud registration, and transformer-based point cloud registration.

\begin{figure*}
  \centering
  \includegraphics[width=1.00\linewidth]{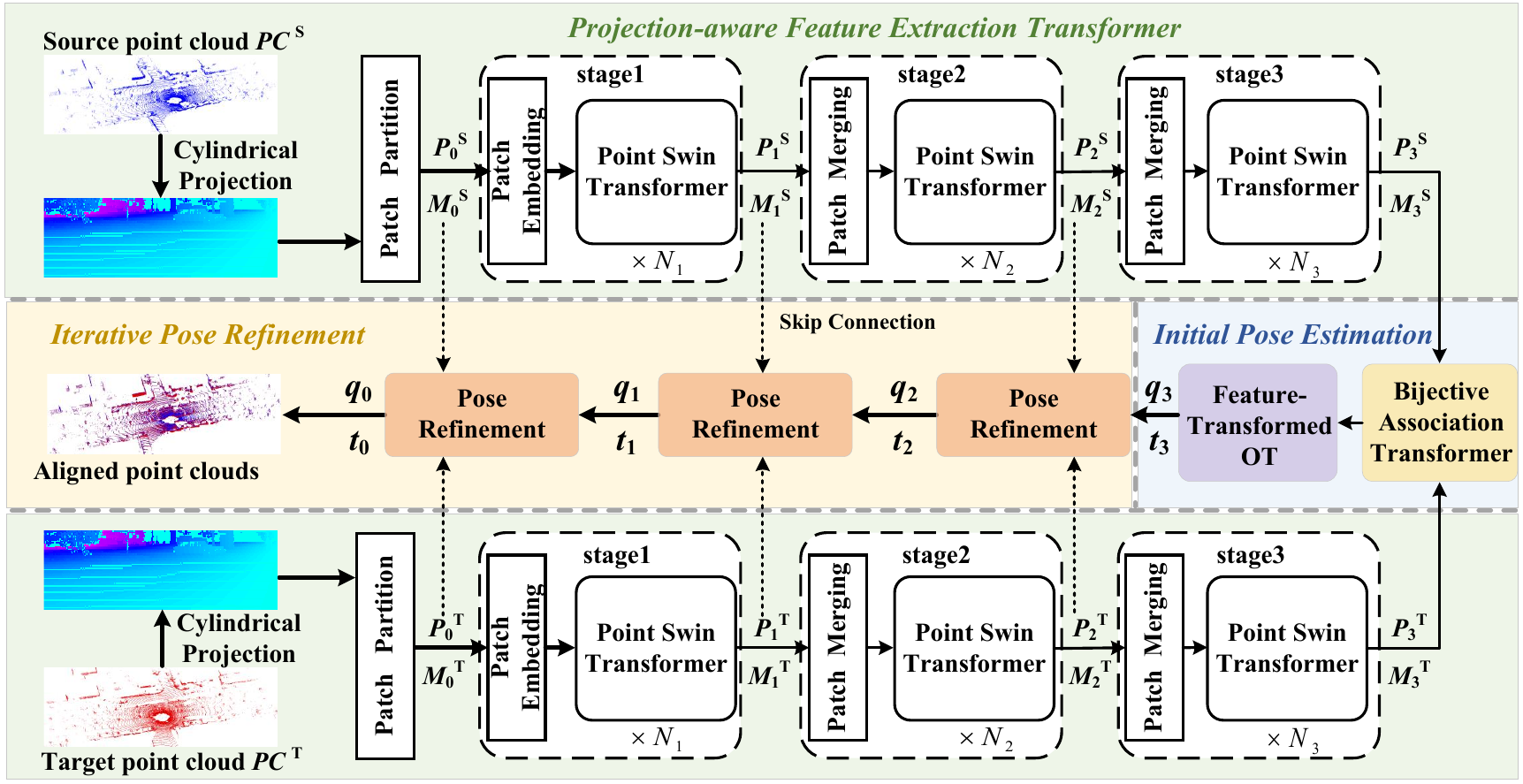}
  \vspace{-6mm}
  \caption{\textbf{The overall architecture of RegFormer++.} We first project point clouds onto a 2D surface and feed its patches into three cascaded feature extraction transformers. For the cross-frame association, we design a Bijective Association Transformer module which includes a cross attention module and an all-to-all point gathering. Finally, transformations are refined iteratively.}
  \label{fig:pipeline}
\end{figure*}

\subsection{Classical Point Cloud Registration} 
Among traditional registration methods, Iterative Closest Point (ICP) \cite{121791} is the best-known solution to align two point cloud frames. They recover the rigid motion by iterative updates based on a least square objective function until convergence. However, ICP is highly sensitive to initialization and often converges to spurious local minima \cite{lu2021hregnet}. Another line of representative works is based on the Normal Distribution Transform (NDT) \cite{biber2003normal}. Two point frames are converted into a normal distribution of multidimensional variables in pre-divided grid cells. The objective of the NDT algorithm is to calculate the transformation matrix that maximizes the sum of their probability densities. Compared to ICP, NDT has lower calculation complexity and more robustness to the initialization value. Additionally, RANdom SAmple Consensus (RANSAC) \cite{fischler1981random} is also commonly adopted to filter outliers in previous registration works, especially feature-based ones. It randomly samples small subsets of correspondences and then evaluates the performance by the percentage of inner points. Some recent works \cite{zhang20233d,yan2025turboreg,wupointtruss} also leverage the maximal clique algorithm \cite{zhang20233d} or K-Truss \cite{wupointtruss} algorithms to solve the registration task.

\subsection{Learning-based Point Cloud Registration} 
Existing deep registration networks can be divided into two categories according to whether they extract explicit correspondences. The first class attempts to establish point correspondences through keypoint detection \cite{lu2019deepvcp,fu2021robust,lu2021hregnet} and learning powerful discriminative descriptors \cite{yew20183dfeat,choy2019fully,bai2020d3feat,ao2021spinnet,deng2018ppfnet}. As a pioneering work, 3DMatch \cite{zeng20173dmatch} extracts local volumetric patch features with a Siamese network. PPFNet \cite{deng2018ppfnet} and its unsupervised version PPF-FoldNet \cite{deng2018ppf} extract global context-aware descriptors using PointNet \cite{qi2017pointnet}. To enlarge the receptive field, FCGF \cite{choy2019fully} computes dense descriptors of whole input point clouds in a forward pass. Subsequent correspondence-based networks \cite{choy2020deep, yang2020teaser,bai2021pointdsc,chen2022deterministic,chen2022sc2} commonly use it to generate putative correspondences. RARE \cite{zheng2025rare} designs a diffusion-based feature extractor to refine any pair-wise correspondences. Regor \cite{zhao2025progressive} reformulates the task as a progressive correspondence regenerator process.

The second class directly estimates transformation in an end-to-end manner \cite{aoki2019pointnetlk,wang2019deep,huang2020feature,yew2020rpm}. Point clouds are aligned with learned soft correspondences or without explicit correspondences. Among these, PointNetLK \cite{aoki2019pointnetlk} is a landmark that extracts global descriptors and estimates transformation with the Lucas-Kanade algorithm \cite{lucas1981iterative}. FMR \cite{huang2020feature} enforces the optimization of registration by minimizing feature-metric projection errors. DeepGMR \cite{yuan2020deepgmr} introduces the Gaussian Mixed Model (GMM) into the registration field, which formulates registration as the minimization of KL divergence between two probability distributions. However, these direct registration methods can not generalize well to large-scale scenes as stated in \cite{huang2021predator,ao2023buffer}. Our method falls into this category and is specially designed for large-scale registration. Motivated by recent scene flow works \cite{jiang2024neurogauss4d,liu2024difflow3d,liu2025difflow3d,zhang2024deflow,zhang2024seflow,jiang20243dsflabelling}, our RegFormer++ requires no keypoint-searching and explicit point-to-point correspondences, that learns implicit cross-frame motion embeddings and directly outputs pose in a single pass.

\subsection{Large-Scale Point Cloud Registration} 
Large-scale registration is less explored in previous works. DeepVCP \cite{lu2019deepvcp} incorporates both local similarity and global geometric constraints in an end-to-end manner. Although it is evaluated on outdoor benchmarks, its keypoint matching is still constrained in local space. With a larger keypoint detection range, HRegNet \cite{lu2021hregnet} introduces bilateral and neighborhood consensus into keypoint features and achieves state-of-the-art. GCL \cite{liu2023density} proposes to extract density-invariant geometric features to register distant outdoor LiDAR point clouds. LS-RegNet \cite{tao2025lsreg} designs a scale attention module. Different from previous works \cite{lu2019deepvcp,lu2021hregnet,wiesmann2022dcpcr} that mostly focus on local descriptors, we address the issue from a more global perspective thanks to the long-range dependencies capturing ability of the transformer.

\subsection{Transformer in the Registration Task} 
Most existing works \cite{wang2019deep,fu2021robust,yu2021cofinet,liu2025frequency,yew2022regtr,qin2022geometric,yu2024riga,du2022swinpa,shao2025mol,ma2024generative,ma2025fine} only treat transformer as a frame association module. Among these, DCP \cite{wang2019deep} first utilizes a vanilla transformer to correlate down-sampled source and target point features. RGM \cite{fu2021robust} proposes a deep graph matching framework, where transformer is employed to dynamically learn the soft edges. REGTR \cite{yew2022regtr} obtains cross-frame information with a cross attention module and then outputs overlap scores and location information. GeoTransformer \cite{qin2022geometric} encodes geometric structure feature, including both pair-wise distances and triplet-wise angles, into its attention calculation. Lepard \cite{li2022lepard} aims to match partial points in rigid and deformable scenes, with transformer for information exchange. RoITr \cite{yu2023rotation} and RIGA \cite{yu2024riga} design rotation-invariant descriptors combining global modeling ability of transformer. PPT \cite{wang2025point} introduces a point tree transformer to enhance efficiency. DFAT \cite{fu2025dual} proposes a dual focus attention to filter irrelevant points. Different from these previous works, our RegFormer++ is fully transformer-based, where transformer is designed for not only frame association but also feature extraction.

\section{RegFormer++}
\subsection{Overall Architecture}

\label{overall}
 The overall architecture of our proposed RegFormer++ is illustrated in Fig.~\ref{fig:pipeline}. Given two point cloud frames: source point cloud $PC^{S}\in\mathbb{R}^{N\times3}$ and target point cloud $PC^{T}\in\mathbb{R}^{M\times3}$, the objective of the registration task is to align them via estimated rotation and translation matrices. To orderly organize raw irregular points, we first project point clouds as pseudo images $P_{0}^{S}$ and $P_{0}^{T}$ in Section~\ref{sec:projection}, and then feed them together with corresponding masks $M_{0}^{S}$ and $M_{0}^{T}$ into the hierarchical feature extraction transformer module. Following prior works \cite{liu2021swin,liu2022video}, we treat each patch of size $4\times 8$ as a token, and then a 2D kernel-based feature embedding layer in Section~\ref{sec:embed} is utilized to efficiently extract point features with an arbitrary dimension denoted by $C$. The patch merging layer of each stage concatenates $2\times 2$ neighbor patches. Then, concatenated features are reduced to half channels, and then fed into a projection-aware Point Swin Transformer in Section~\ref{pst}. For associating point clouds and reducing mismatches, a Bijective Association Transformer (BAT) in Section~\ref{bat} is employed to generate initial motion embeddings. Finally, the quaternion vector $q_{3}\in\mathbb{R}^{4}$ and translation vector $t_{3}\in\mathbb{R}^{3}$ are estimated from motion embeddings, and then refined iteratively as in Section~\ref{sec:transform}.
 
\begin{figure}
 \centering
 \includegraphics[width=1.0\linewidth]{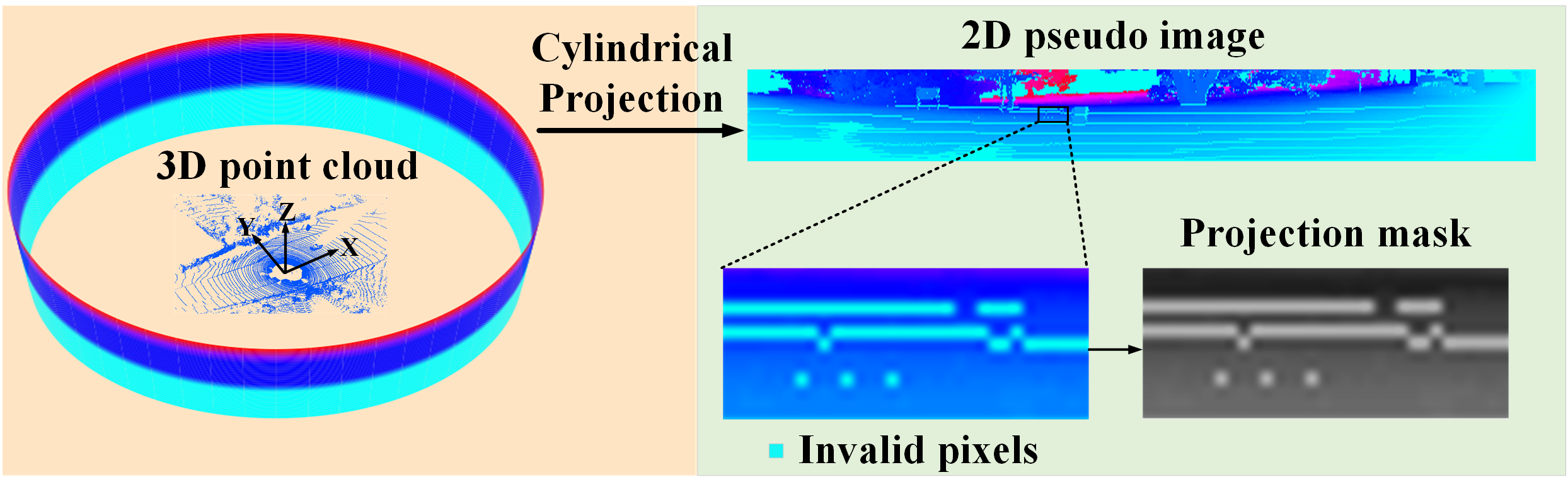}
 \vspace{-6mm}
 \caption{Cylindrical projection. We project 3D point clouds onto a 2D surface and fill each pixel with its raw $x,y,z$ coordinates. A projection mask is also proposed to remove invalid positions.}
 \label{fig:pro}
\end{figure}
\subsection{Cylindrical Projection}
\label{sec:projection}
 Cylindrical projection has been widely adopted in recent 3D point cloud learning networks, ranging from semantic segmentation \cite{ando2023rangevit,kong2023rethinking}, point cloud generation \cite{zyrianov2022learning}, motion prediction \cite{luo2023pcpnet}, and LiDAR odometry \cite{li2019net,li2020dmlo,nicolai2016deep}  fields. To convert the original sparse and irregular point clouds into a structured representation, they project points onto a cylindrical surface and then utilize the 2D convolutional network for high processing efficiency. In this paper, we also adopt cylindrical projection, but leverage a 2D transformer maintaining 3D geometric information to extract global features with robustness to outliers proved in \cite{yu2024riga}.


Specifically, following the line scanning characteristic of the LiDAR sensor, point clouds are projected onto a cylindrical surface, obeying the original proximity relationship of raw points. Each point has a corresponding 2D pixel position on projected pseudo images as:
\begin{equation}
    \label{eq:projection1}
     u = arctan 2(y/x)/\Delta\theta,
\end{equation}

\begin{equation}
\label{eq:projection2}
     v = arcsin (z/\sqrt{x^2 + y^2 +z^2})/\Delta\phi,
\end{equation}
where $x,y,z$ represent the raw 3D coordinates of points and $u,v$ are corresponding 2D pixel positions. $\Delta\theta$ and $\Delta\phi$ are horizontal and vertical resolutions of the LiDAR sensor, respectively. To make full use of the geometric information of the raw 3D points, we fill each pixel position with its raw coordinates $x,y,z$. Pseudo images of size $H\times W\times 3$ in Fig.~\ref{fig:pro} will be input to the feature extraction transformer.

\subsection{Kernel-based Efficient Patch Embedding}
\label{sec:embed}
According to the scanning characteristic of mechanical LiDAR sensors, points are usually closer in 3D space when they are closer in the projected 2D space. To make full use of the original geometric features and also keep high efficiency, a 2D kernel-based CUDA operator from \cite{wang2022efficient} is introduced here to gather the features of the neighbor patch. 

\begin{figure}
  \centering
  \includegraphics[width=1.00\linewidth]{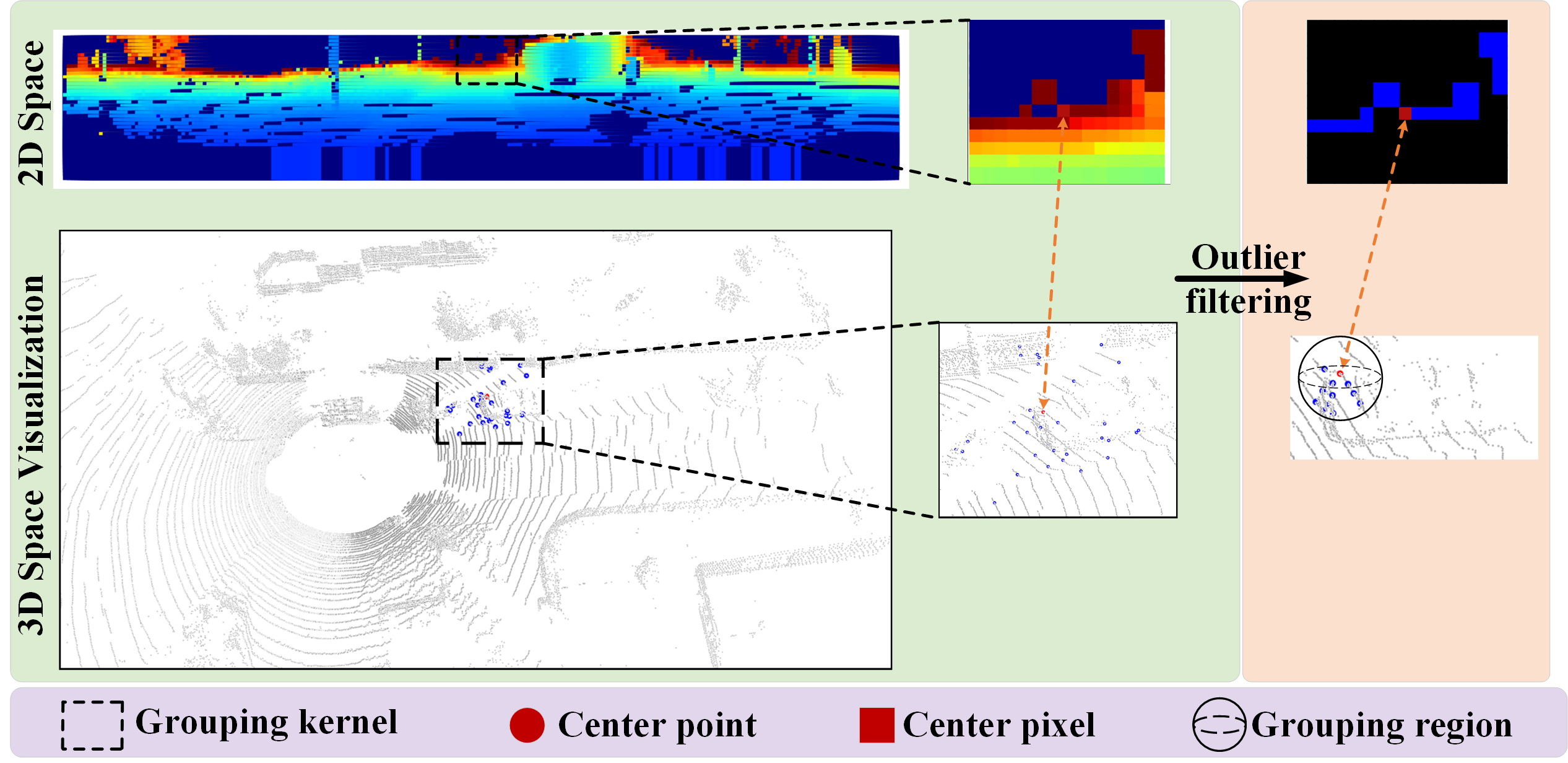}
  \vspace{-6mm}
  \caption{Visualization of kernel-based efficient patch embedding. For each center point, its surrounding neighbor points within a predefined 2D kernel are indexed and grouped. Then, outliers that are harmful to registration are filtered. Above grouping and filtering in 2D space actually happen in 3D space based on the cylindrical projection process. }
  \label{fig:kernel}
  
\end{figure}

\vspace{3pt}
\noindent\textbf{CUDA-based Feature Gathering.} As in Fig. \ref{fig:kernel}, a series of centers are first predefined according to the stride similar to 2D convolutional networks. Neighbor searching around each center point is then conducted within the 2D kernel. During the searching process, far-away points that are harmful to feature extraction are also filtered. Notably, above neighbor searching, grouping, and filtering are all operated on the projected 2D plane with the parallel-processing CUDA operator, but its corresponding process actually happens in the 3D space, because we have filled projected 2D positions with corresponding 3D coordinates. Therefore, this operation can not only retain raw 3D geometric information but also keep high efficiency with our designed 2D operators \cite{peng2023delflow}. 

\vspace{3pt}
\noindent\textbf{Advantages Compared to Prior Transformers.} Some recent literature like GeoTransformer \cite{qin2022geometric} has also attempted to take advantage of raw 3D geometric features in transformer by additionally introducing pair-wise distance and triplet-wise angle features inside the attention block. However, their geometric module is performed in each feature extraction layer, and conducted directly in 3D space, which extremely increases the computation burdens as in Table~\ref{table:ransac}. Compared to GeoTransformer, our feature embedding with a 2D CUDA-based operator is extremely effective, only performed once in the first patch embedding layer. Then, only 2D patch merging and attention layers are utilized to extract more high-level features. Experiments show that our designed architecture can effectively gather distinctive features with higher registration accuracy, and also achieve high efficiency with only $6.3\%$ of the inference time compared to GeoTransformer.

\subsection{Point Swin Transformer} 
\label{pst}
It is infeasible to leverage the vanilla transformer with quadratic complexity for modeling outdoor LiDAR points. Compared with images, the scale of outdoor LiDAR points is surprisingly larger, and thus requires substantial tokens for representation. Inspired by Swin Transformer \cite{liu2021swin}, we revisit window attention in 3D point transformer for linear complexity. Thanks to the global understanding ability of transformer, our network can effectively learn to identify dynamic motions and the location of occluded objects in the other frame, as they introduce inconsistent motions \cite{yu2024riga}. To simplify the formulation, we only expand the descriptions for $PC^{S}$ in this section, and the same goes for $PC^{T}$.

\vspace{3pt}
\noindent\textbf{Projection Masks.} It is non-trivial to extend the 2D window-based attention to the pseudo images generated from 3D points. Point cloud, especially in outdoor scenes, is extremely sparse. Thus, projected pseudo images are filled with invalid blank pixels. The attention calculation of these pixels is meaningless as they have no corresponding raw points. Inspired by \cite{cheng2021per,liu2023translo}, a projection-aware binary mask $M_{l}^{S}$ is proposed here, which represents whether each pixel is invalid in Fig.~\ref{fig:pro}:
\begin{equation}
M_{l}^{S}=
\begin{cases}
-\infty,& \text{ $ x = 0,y = 0,z = 0 $ }, \\
0,& \text{ otherwise },
\end{cases}
\end{equation}
where $x,y,z$ are point coordinates filled into pseudo images. The projection mask $M_{l}^{S}$ of size $H\times W\times 1$ is pixel-corresponding to the pseudo image and together downsampled (in a stride-based manner) in each stage as in Fig.~\ref{fig:pipeline}. $l$ denotes the stage number. We assign zero to valid pixels and a big negative number to invalid ones. In this way, invalid pixels would then be filtered through the softmax operation in the following attention blocks.

\begin{figure*}
 \centering
 \includegraphics[width=1.0\linewidth]{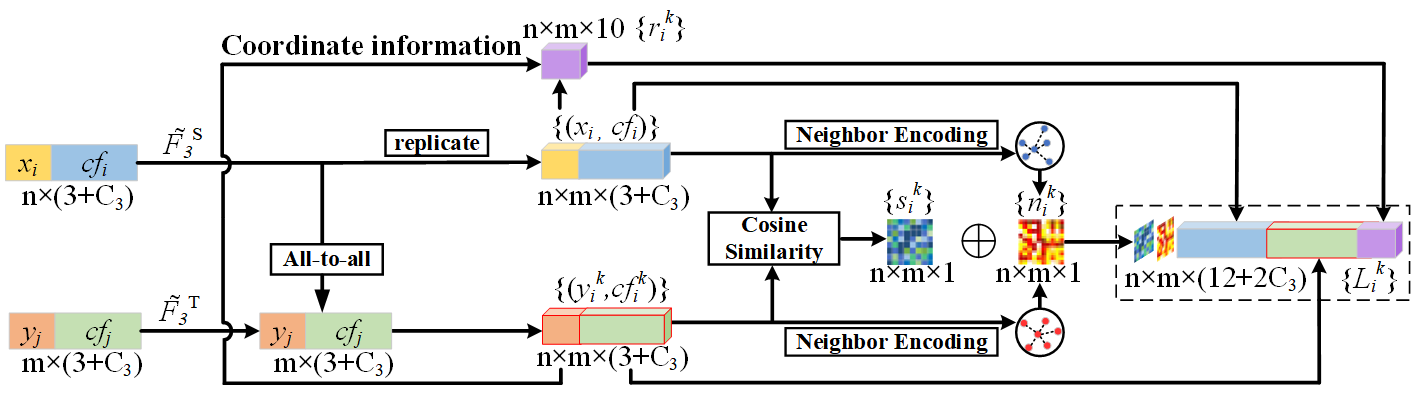}
 \vspace{-7mm}
 \caption{All-to-all point gathering. After the cross-attention mechanism leveraged for preliminary information exchange between two frames, geometric characteristics of conditioned features ${\tilde{F}}^{S}_{3}, {\tilde{F}}^{T}_{3}$ are fully considered to generate the initial motion embeddings. The initial embeddings include coordinate information, point and neighbor similarity features, and content features.}
 \label{fig:bat}
\end{figure*}

\vspace{3pt}\noindent\textbf{Point W-MSA and Point SW-MSA.} For stage $l$, point feature $P_{l}^{S}$ ($H_{l}\times W_{l}\times C_{l}$) and its corresponding mask $M_{l}^{S}$ ($H_{l}\times W_{l}\times 1$) are fed into Point Window-based Multi-Head Self Attention (Point W-MSA) as: 
\begin{equation}
    \label{eq:WMSA}
    \text{W-MSA}(P_{l}^{S}) = (Head_1 \oplus \cdots \oplus Head_H)W^O, 
\end{equation}
\begin{align}
      \label{eq:head}
               Head_h 
              &= Attention({Q^h}, {K^h}, {V^h}) \notag\\ 
              &\vspace{-10pt}= \text{softmax}(\frac{{Q^h}{K^h}}{\sqrt{d_{head}}}+ M_{l}^{S}+ Bias)V^h,
\end{align}
where $Head_h$ represents the output of the $h-th$ head. $\oplus$ denotes the concatenation operations. $Bias$ is the relative position encoding \cite{raffel2020exploring}. $Q^h=P_{l}^{S}{W_h^Q}$, $K^h=P_{l}^{S}{W_h^K}$, ${V^h}=P_{l}^{S}{W_h^V}$, in which $W_h^Q\in{\mathbb{R}^{ C_{l} \times C_{head}}}$, $W_h^K\in{\mathbb{R}^{ C_{l} \times C_{head}}}$, $W_h^V\in{\mathbb{R}^{ C_{l} \times C_{head}}}$, $W^O\in{\mathbb{R}^{H C_{head} \times C_{l}}}$ are learnable projection functions. Then, output features are spatially shifted \cite{liu2021swin} and calculated attention again in the following Point Shift Window-based Multi-head Self Attention (Point SW-MSA), to increase the cross-window interaction.

\vspace{3pt}\noindent\textbf{Point Swin Transformer Blocks.} Overall, one complete transformer stage can be described as:
\begin{align}
    \label{transformer}
    &{\hat{P}}_{l}^{S} = \text{PW-MSA}(\text{LN}(P_{l}^{S})) + P_{l}^{S}, \notag\\
    &{Z_{l}^{S}} = \text{MLP}(\text{LN}({\hat{P}}_{l}^{S})) + {\hat{P}}_{l}^{S}, \notag\\
    &{\hat{P}}_{l+1}^{S} = \text{PSW-MSA}(\text{LN}({Z_{l}^{S}})) + {Z_{l}^{S}}, \notag\\
    &{P_{l+1}^{S}} = \text{MLP}(\text{LN}({\hat{P}}_{l+1}^{S})) + {\hat{P}}_{l+1}^{S},
\end{align}
where ${{P}}_{l+1}^{S}$ is the output feature of stage $l$ and then delivered into the stage $l+1$ as input features.

\subsection{Bijective Association Transformer}
\label{bat}
After global features are hierarchically extracted by our Point Swin Transformer, the key issue is how to accurately match source and target point clouds through their downsampled features. The most common method is to search for nearest neighbors (NN). However, this distance-dependent strategy is not effective enough for large-scale registration, as point correspondences may be too far away and have limited shared ranges with low overlap, which leads to numerous mismatches \cite{lu2021hregnet}. To solve this problem, we propose a Bijective Association Transformer block (BAT) on the coarsest layer. A cross-attention layer first exchanges preliminary contextual information between two frames. Then, an all-to-all point-gathering strategy is proposed to further guarantee reliable location embeddings and reduce mismatches.

\vspace{3pt}\noindent\textbf{Cross Attention for Information Exchange.} As depicted in Fig.~\ref{fig:pipeline}, down-sampled source and target point features of stage 3 are first fed into a cross-attention layer with linear complexity, roughly associating with each other. Cross attention can introduce certain similarities of two point cloud frames by calculating attention weights and updating features with the awareness of point location in the other frame \cite{yew2022regtr}. Concretely, source and target point features are first resized as ${P^S_3}\in\mathbb{R}^{n\times C_{3}}$ and ${P^T_3}\in\mathbb{R}^{m\times C_{3}}$, which are inputs of the cross-attention block. The output features ${\tilde{P}}^{S}_3$ for the source point cloud can be written as:
\begin{equation}
    \label{eq:cross}
   {\tilde{F}}^{S}_{3}
   = Attention({P^{S}_{3}{W^Q}}, {P^{T}_{3}{W^K}}, {P^{T}_{3}{W^V}}),
\end{equation}
where $W^Q,W^K, W^V$ are projected functions. $Attention$ denotes the same window-based attention with spatial shift as in Section~\ref{pst}. When $PC^{S}$ serves as $query$, $PC^{T}$ would be projected as $key$ and $value$, and vice versa.

\vspace{3pt}\noindent\textbf{All-to-All Point Gathering.} The coarsest layer obviously gathers more information and has a larger receptive field, which is reliable to match two frames. Thus, on the bottom layer of our RegFormer++, each point in ${\tilde{F}}^{S}_{3}$  is associated with all points in ${\tilde{F}}^{T}_{3}$, rather than select $k$ nearest neighbor points ($k$NN), to generate reliable motion embeddings. Specifically, in Fig. \ref{fig:bat}, each point in ${PC^S}=\{(x_i,cf_i)|{x_i}\in{X^S},{cf_i}\in{\tilde{F}}^{S}_{3}, i=1,\cdots,n\}$ correlates with all $m$ points in ${PC^T}=\{(y_j,cf_j)|{y_j}\in{Y^T},{cf_j}\in{\tilde{F}}^{T}_{3}, j=1,\cdots,m\}$, forming an association cluster $\{(x_{i},y^{k}_{i})|k=1,\cdots,m\}$. Then, the relative 3D Euclidean spatial information $\{r^{k}_{i}\}$ is calculated as:
\begin{equation}
    \label{eq:dis}
    r^{k}_{i} = {x_i} \oplus {y^{k}_{i}} \oplus (x_i - y^{k}_{i}) \oplus \Vert x_i - y^{k}_{i} \Vert_{2},
\end{equation}
where $\left\|  \cdot  \right\|_2$ indicates the $L_2$ Norm. 

The cosine similarity of grouped features is also introduced as:
\begin{equation}
    \label{eq:feat}
    s^{k}_{i} = \frac{<cf_i,cf_{i}^{k}>}{\Vert cf_i\Vert_{2} \Vert cf^{k}_{i}\Vert_{2}},
\end{equation}
where $< ,>$ denotes the inner product. This step will output a $n\times m\times 1$ point similarity feature $s^{k}_{i}$. Also, a neighbor similarity $n^{k}_{i}$ \cite{lu2021hregnet} with the shape of $n\times m\times 1$ is generated similarly.


\begin{algorithm}[tb]
\caption{Feature-Transformed Optimal Transport}
\label{alg:ot}
\textbf{Input}: source point coordinates and features ${X}_{3},{\tilde{F}}^{S}_{3}$, target point coordinates and features ${Y}_{3},{\tilde{F}}^{T}_{3}$, motion embeddings FE.\\
\textbf{Parameter: SINKHORN}{(${\tilde{F}}^{S}_{3}, {\tilde{F}}^{T}_{3}, {X}_{3}, {Y}_{3}, $FE$, \epsilon, \gamma$, I)}.\\
\textbf{Output}: transformed point features {FE}$_{trans}$, transformed flow $SF$.

\begin{algorithmic}[1]
\State Compute cost matrix between features:  
 $C \Leftarrow f({\tilde{F}}^{S}_{3}, {\tilde{F}}^{T}_{3})$ 
\State $K \Leftarrow \exp(-C / \epsilon)$ \{$\epsilon$ is the regularization term.\}
\State Initialize scaling vector, with $n$ as rows of ${\tilde{F}}^{S}_{3}$: 
        $u^{0} \Leftarrow \mathbf{1}_{n}$ 
\For{$i \Leftarrow 0$ to I-1} 
    \State$v^{i+1} \Leftarrow \gamma / (K^\top u^{i})$
    \State$u^{i+1} \Leftarrow \gamma / (K v^{i+1})$
\EndFor
\State Compute the transport matrix: $T \Leftarrow u^{I} \otimes v^{I} \otimes K$
\State Row sum of $T$: $R_s \Leftarrow \text{sum}(T, \text{axis}=-1, \text{keepdim}=\text{True})$ 
\State Compute OT flow: $SF \Leftarrow (T \times X_3) / (R_s + 10^{-8}) - Y_3$ 
\State Normalize transport matrix: $NT \Leftarrow T / \text{sum}(T, \text{axis}=1)$
\State Transform motion embeddings FE using $NT$: \\{FE}$_{trans}$$ \Leftarrow NT \times$ FE \\
\Return{{FE}$_{trans}, SF$}
\end{algorithmic}
\end{algorithm}
Then, we concatenate the above spatial features, point features, and similarity embeddings together, and then utilize a 3-layer shared MLP on them:
\begin{equation}
    \label{eq:cat}
    L^{k}_{i} = \text{MLP}(cf_{i}\oplus cf^{k}_{i}\oplus r^{k}_{i}\oplus n^{k}_{i}\oplus s^{k}_{i}).
\end{equation}

Finally, the initial flow embedding can be represented by the attentive encoding of concatenated features as:
\begin{equation}
    \label{eq:flow1}
    fe_{i} = \sum\limits_{k =1}^{m} {L^{k}_{i}} \odot{\mathop{\text{softmax}}\limits_{k =1,\cdots,m}{(L^{k}_{i})}},
\end{equation}
where the output motion embedding $fe_{i}$ is a weighted sum of $ L^{k}_{i}$ by the softmax function. $\odot$ represents dot product.

\subsection{Feature-Transformed Optimal Transport} 
After obtaining the motion embedding $fe_{i}$, our former work RegFormer will directly regress the rotation and translation from the feature. However, we find that this direct regression is fragile to network training parameters because there is no explicit correspondence across frames. To improve the motion stability, we further design an optimal transport module to transform motion features. The optimal transport section is established on the mathematical cost matrix between point features from two frames, naturally robust to different training parameters. As in Algorithm~\ref{alg:ot}, we design a feature-transformation sinkhorn algorithm to strengthen the above motion embeddings $FE =\{fe_{i},i=1,\cdots,n\}$. Specifically, we first compute a cost matrix representing the optimal transformation from source point features ${\tilde{F}}^{S}_{3}$ to the target ones ${\tilde{F}}^{T}_{3}$. After $I$ iterations, we obtain the transformed motion features $FE_{trans}$ by transforming motion embeddings $FE$ with the normalized transport matrix $NT$. Also, a transformed residual flow $SF$ is generated by applying a transport matrix to the point coordinates. Finally, the output motion embeddings $ME$ are generated by:
\begin{equation}
    \label{eq:cat}
    ME = FE \oplus FE_{trans}\oplus SF.
\end{equation}

\begin{table*}[t]
	\centering
	\footnotesize
    \caption{Detailed network settings in transformer layers. $N$ denotes the number of attention blocks. MLP width indicates the expansion layer of MLP in feed-forward networks. }
    \vspace{-5mm}
	\begin{center}
		\setlength{\tabcolsep}{2mm}
		{
        \resizebox{1.0\textwidth}{!}
		{
			\begin{tabular}{@{}ccccccccc@{}}
						\toprule
						Module & \multicolumn{2}{c}{Layer type} &$N$ & Sample rate & Channels &MLP width&Heads&Dropout\\ \midrule
						&\multicolumn{2}{c}{Point Swin Transformer (stage 1)} & 2 & $[H/4\times W/8]$ & 16 &[16,64,16]&2&0.1\\
						&\multicolumn{2}{c}{Point Swin Transformer (stage 2)} & 2 & $[H/8\times W/16]$  & 32&[32,128,32]&4&0.1\\
						\multirow{-3}{*}{\begin{tabular}[c]{@{}c@{}}Feature Extraction\\Transformer \end{tabular}}&\multicolumn{2}{c}{Point Swin Transformer (stage 3)} & 6 & $[H/16\times W/32]$& 64 &[64,256,64]&8&0.1\\
						
						\cline{1-9}\noalign{\smallskip}

                        &\multicolumn{2}{c}{self-attention layer} & 6 & --- & 64 &[64,256,64]&8&0.1\\
						
						  \multirow{-2}{*}{\begin{tabular}[c]{@{}c@{}}Bijective Association\\Transformer \end{tabular}}&\multicolumn{2}{c}{cross-attention layer} & 6 & --- & 64 &[64,256,64]&8&0.1\\
						
						\bottomrule
						
				\end{tabular}
		}}
	\end{center}
    
    \label{table:parameters1}
\end{table*}

\begin{table*}[t]
	\centering
	\footnotesize
    \caption{Detailed network parameters in the initial pose estimation and iterative refinement layers. $K$ points are selected in the $K$ Nearest Neighbors (KNN) of the all-to-all point gathering layer, set upconv layer, and attentive cost volume layer. MLP width means the number of output channels for each layer of MLP.}
    \vspace{-5mm}
	\begin{center}
		\setlength{\tabcolsep}{2mm}
		{
         \resizebox{1.0\textwidth}{!}
		{
			\begin{tabular}{@{}cccccc@{}}
						\toprule
						Module & \multicolumn{2}{c}{Layer type} &$K$ & Sample rate & MLP width \\ \midrule

						&\multicolumn{2}{c}{All-to-all point gathering}  &  {112}  & 1 & {[128,64,64], [128,64]} \\
						
						&\multicolumn{2}{c}{Shared MLP for $ME_{3}$}    & ---   & 1 & [128,64] \\
						
						&\multicolumn{2}{c}{$FC_{1}$ for $q_3$, $FC_2$ for $t_3$}  & --- & 1 & [4], [3] \\
						
						\cline{1-6}\noalign{\smallskip}
						\multirow{-5.5}{*}{\begin{tabular}[c]{@{}c@{}}Estimation of \\Initial Transformation \end{tabular}}	
                        
						&&Attentive cost volume  & 4, 6  & 1  & [128,64,64], [128,64] \\
						&&Set upconv& 8   & $2\times2$ & [128,64], [64] \\	
						&&Shared MLP for $ME_2$  & --- &  1  & [128,64] \\
						
						&\multirow{-4}{*}{\begin{tabular}[c]{@{}c@{}}Pose Refinement for $q_2$,$t_2$ \end{tabular}}  
						&$FC_{1}$ for $q_2$, $FC_{2}$ for $t_2$  & --- & 1 & [4], [3] \\
						
						\cline{2-6}\noalign{\smallskip}					
						&&Attentive cost volume   & 4, 6  & 1  & [128,64,64], [128,64] \\
						&&Set upconv & 8   & $2\times2$ & [128,64], [64] \\	
						&&Shared MLP for $ME_1$& --- &  1  & [128,64] \\

						&\multirow{-4}{*}{\begin{tabular}[c]{@{}c@{}}Pose Refinement for $q_1$,$t_1$\end{tabular}}  
						&$FC_{1}$ for $q_1$, $FC_{2}$ for $t_1$  & --- & 1 & [4], [3] \\
						
						\cline{2-6}\noalign{\smallskip}	
						
						&&Attentive cost volume  & 4, 10  & 1  & [128,64,64], [128,64] \\
						&&Set upconv & 8   & $4\times8$ & [128,64], [64] \\	
						&&Shared MLP for $ME_0$  & --- &  1  & [128,64] \\

						&\multirow{-4}{*}{\begin{tabular}[c]{@{}c@{}}Pose Refinement for $q_0$,$t_0$\end{tabular}}  
						&$FC_{1}$ for $q_0$, $FC_{2}$ for $t_0$  & --- & 1 & [4], [3] \\
						
						\bottomrule
						\multirow{-14.5}{*}{\begin{tabular}[c]{@{}c@{}}Iterative Pose Refinement \\with PWC Structure \cite{wang2021pwclo}\end{tabular}} 
				\end{tabular}
		}}
	\end{center}
	\label{table:parameters2}
\end{table*}


\subsection{Estimation of the Rigid Transformation} 
\label{sec:transform}
The initial transformation is estimated from motion embeddings $ME=\{me_{i},i=1,\cdots,n\}$ together with down-sampled source point features $F_3^S$ in layer 3 as:
\begin{equation}
    \label{eq:pose}
    W = \text{softmax}(\text{MLP}(ME \oplus {F_3^S})),
\end{equation}
where $W=\{w_i|w_i\in \mathbb{R} ^{C_{3}}\}$ are attention weights. Then, the quaternion $q_{3}\in\mathbb{R}^{4}$ and translation vector $t_{3}\in\mathbb{R}^{3}$ can be generated separately from weighting and sum operations followed by a fully connected layer \cite{wang2021pwclo}:
\begin{equation}
    \label{eq:pose1}
   q^3 = \frac{FC_1(\sum\limits_{i=1}^{n} me_{i}\odot w_i)}{|FC_1(\sum\limits_{i=1}^{n} me_{i}\odot w_i)|},
\end{equation}
\begin{equation}
   \label{eq:pose2}
   t^3 = FC_2(\sum\limits_{i=1}^{n} me_{i}\odot w_i),
\end{equation}
where $FC_1$ and $FC_2$ denote two fully connected layers.

Nonetheless, the initially estimated transformation is not precise enough due to the sparsity of the coarsest layer. Thus, we iteratively refine it on upper layers with the PWC structure \cite{sun2018pwc, wang2021pwclo} to generate residual transformation $\Delta q^l$ and $\Delta t^l$. Refinement in the $l-th$ layer can be indicated as:
\begin{equation}
   \label{eq:refine}
   q^l = \Delta q^l q^{l+1},
\end{equation}
\begin{equation}
   \label{eq:refine1}
   [0,t^l] = \Delta q^l [0,t^{l+1}] ({\Delta q^l})^{-1}+[0,\Delta t^l].
\end{equation}

\subsection{Loss Function}
Our network outputs transformation parameters from four layers and adopts a multi-scale supervised approach:
${L}=\alpha^l  \mathcal{L}^l$. $\alpha^l$ indicates weights of layer $l$. $\mathcal{L}^l$ denotes the loss function of the $l-th$ layer, which is calculated as:
\begin{equation}
   \mathcal{L}^l= \mathcal{L}_{trans}^l exp(-k_t) + k_{t} + \mathcal{L}_{rot}^l exp(-k_r) + k_{r},
\end{equation}
where $k_{t}$ and $k_{r}$ are two learnable parameters, which can uniform differences in the unit and scale between quaternion and translation vectors \cite{li2019net}. $\mathcal{L}_{trans}^l$ and $\mathcal{L}_{rot}^l$ can be calculated as:
\begin{equation}
   \label{eq:loss}
   \mathcal{L}_{trans}^l= \Vert t^l-\hat{t}^l\Vert,
\end{equation}
\begin{equation}
   \label{eq:loss1}
   \mathcal{L}_{rot}^l= \Vert \frac{q^l}{\Vert q^l\Vert}-\hat{q}^l\Vert_2,
\end{equation} 
where $\left\|  \cdot  \right\|$ indicates the $L_1$ Norm. $q^l, t^l$ and $\hat{q}^l,\hat{t}^l$ are estimated and ground truth transformations, respectively.

\section{Experiment}
We evaluate our RegFormer++ on three large-scale point cloud registration datasets, namely KITTI \cite{geiger2012we,geiger2013vision}, NuScenes \cite{caesar2020nuscenes}, and Argoverse \cite{chang2019argoverse}. Moreover, we generalize our pre-trained RegFormer++ to the unseen ETH dataset to demonstrate the strong generalization capability. For real-time application, we leverage our registration method on the LiDAR odometry task, which achieves highly competitive performance. Ablation studies are conducted for each designed component of our network to demonstrate their effectiveness. Extensive experiments demonstrate that our methods can achieve state-of-the-art registration accuracy and also guarantee high efficiency.

\subsection{Experiment Settings}
\textbf{Implement Details.} Unlike previous registration methods \cite{yew20183dfeat,choy2019fully,bai2020d3feat,yu2021cofinet,huang2021predator,ao2021spinnet} which voxelize input points, we directly input all LiDAR points without down-sampling. The resolution of projected pseudo images are set in line with corresponding beams of LiDAR sensor as 64 ($H$)$\times$1792 ($W$) for KITTI and 32 ($H$)$\times$1792 ($W$) for NuScenes \& Argoverse. Window size is set as 4. Experiments are conducted on a single NVIDIA RTX3090 GPU with PyTorch 1.12.0. The Adam optimizer is adopted with $\beta_{1}$ = 0.9, $\beta_{2}$ = 0.999. The initial learning rate is 0.001 and exponentially decays every 200000 steps until 0.00001. The batch size is set as 16. The hyperparameter $\alpha^l$ in the loss function is set to 1.6, 0.8, 0.4, and 0.2 for four layers. Initial values of learnable parameters $k_{t}$ and $k_{r}$ are set as 0.0 and -2.5 respectively.

\vspace{3pt}\noindent\textbf{Network Parameters.} We list the detailed network parameters for our designed transformer layers in Table~\ref{table:parameters1} and settings of pose estimation and refinement layers in Table~\ref{table:parameters2}. For the feature extraction transformer, we establish three successive down-sampling stages, where the down-sampling rates are set as $[4,2,2]$ for the height and $[8,2,2]$ for the width. Point feature channels are projected as 16 in the 2D kernel-based feature embedding layer. And then channels double after each patch merging layer. We set the head numbers in feature extraction transformer layers as 2,4,8 for three stages. Bijective Association Transformer (BAT) is composed of six layers. Each layer consists of a self attention module for intra-frame information interaction and a cross attention layer for inter-frame information exchange. 

For the initial pose estimation, each point in one frame is correlated with all $K=112$ points in the other frame, as in Table~\ref{table:parameters2}. Then, the PWC (Pyramid, Warp, and Cost-volume) refinement structure \cite{sun2018pwc} is applied to recover the precise transformation. First, the source point cloud is warped by the transformation from the upper layers. In this case, the warped source point cloud is closer to the target one. Then, we calculate the attentive cost volume \cite{wang2021hierarchical} between warped source point cloud features and original target point cloud features. Finally, features from the cost volume layer together with up-sampled features will go through a shared MLP for generating residual motion embeddings. The detailed parameters are provided in Table~\ref{table:parameters2}.

\vspace{3pt}\noindent\textbf{Evaluation Metrics.} We follow protocols of DGR \cite{choy2020deep} to evaluate our RegFormer++ with three metrics: (1) Relative Translation Error (RTE): the Euclidean distance between predicted and ground-truth translation vectors. (2) Relative Rotation Error (RRE): the geodesic distance between estimated and ground-truth rotation parameters. (3) Registration Recall (RR): the average success ratio of the registration. Registration is successful when RRE and RTE are within a certain threshold. RRE and RTE can be calculated as:
\begin{equation}
   \label{eq:rre}
   RRE= arccos \frac {Tr({R_{pred}}^{T} R_{gt}-1)}{2},
\end{equation}
\begin{equation}
   \label{eq:rte}
   RTE= \Vert t_{gt}-t_{pred} \Vert_{2},
\end{equation}
where $R_{pred}$ and $t_{pred}$ are predicted rotation and translation vectors. $R_{gt}$ and $t_{gt}$ are the ground truth ones. Note that failed registrations can lead to unreliable error metrics, so we only calculate RRE and RTE for successful registrations following \cite{choy2020deep, choy2019fully, bai2020d3feat}. 

\setlength{\tabcolsep}{0.8mm}
\begin{table}[t]
    
	\centering
	\footnotesize
    \caption{Comparison results with state-of-the-arts on the KITTI dataset. Input point pairs are 10-frame away prepared by \cite{lu2021hregnet}. The best performance is highlighted in bold. Registration Recall (RR) is defined as the success ratio where RRE$<$ 5\degree and RTE$<$ 2m. `NT' means normalized time per thousand points.}	
    \vspace{-5mm}
	\begin{center}
		\resizebox{1.0\columnwidth}{!}
		{
			\begin{tabular}{l|cc|cc|c|c|c}
				\toprule
				&    \multicolumn{2}{c|}{RTE(m)} &\multicolumn{2}{c|}{RRE(\degree)} &  & & \\ 
				
				
				\multirow{-2}{*}{\begin{tabular}[c]{@{}c@{}}Method \end{tabular}}
				&  AVG  & STD   & AVG    & STD        &\multirow{-2}{*}{\begin{tabular}[c]{@{}c@{}}RR(\%) \end{tabular}}   & \multirow{-2}{*}{\begin{tabular}[c]{@{}c@{}}Time(ms)/Points \end{tabular}} &\multirow{-2}{*}{\begin{tabular}[c]{@{}c@{}}NT(ms) \end{tabular}}\\
				\hline\hline
				\noalign{\smallskip}

                FGR \cite{zhou2016fast}
                &0.93 &0.59
                &0.96 &0.81
                &39.4\%&506.1/11445&44.22\\

                RANSAC \cite{fischler1981random}
                &0.13 &0.07
                &0.54 &0.40
                &91.9\% &549.6/16384&33.54\\

                DCP \cite{wang2019deep}    
				&1.03 & 0.51 
				&2.07 & 1.19 
				&47.3\% & 46.4/1024&45.31\\ 

                IDAM \cite{li2020iterative}    
				&0.66 & 0.48 
				&1.06 & 0.94 
				&70.9\% & 33.4/4096&8.15\\ 

                FMR \cite{huang2020feature}    
				&0.66 & 0.42 
				&1.49 & 0.85 
				&90.6\% & 85.5/12000 &7.13\\ 

                DGR \cite{choy2020deep}    
				&0.32 & 0.32 
				&0.37 & 0.30 
				&98.7\% & 1496.6/16384&91.35\\ 
	
				HRegNet \cite{lu2021hregnet}      
				&0.12 & 0.13
				&0.29 & 0.25
				&99.7\%    & 106.2/16384&6.48\\

                 PTT \cite{wang2025point}
                &0.06  &-
                &0.23 &-
				&99.8\% &-&-\\

                LS-RegNet \cite{tao2025lsreg}
                 &0.06  &0.07
                &0.20 &0.17
				&99.9\% &92.5/16384&5.65\\

                \cline{1-8}\noalign{\smallskip}

                RegFormer \cite{Liu_2023_ICCV}
                &0.08 &0.11
                & 0.23 &0.21
                &99.8\% &\bf{98.3/120000}&\bf{0.82}\\

                RegFormer++
                &\bf0.04 &\bf0.06
                &\bf 0.14 &\bf0.13
                &\bf{100\%} &136.2/120000&1.14\\
                \bottomrule
			\end{tabular}
		}
	\end{center}
	\label{table:2m}

\end{table}

\setlength{\tabcolsep}{2mm}
\begin{table}[t]
	\centering
	\footnotesize
    \caption{Comparison results with methods using traditional estimators. Methods are classified into two main categories: RANSAC-based and Graph-based. $^{\dagger}$ indicates methods using optimization algorithm. Other methods including ours are learning-based. Input point pairs are at least 10m away as in \cite{qin2023geotransformer} The best performances are respectively highlighted in bold. RR is defined as the success ratio where RRE$<$ 5\degree and RTE$<$ 2m. }
    \vspace{-5mm}
	\begin{center}
		\resizebox{1.0\columnwidth}{!}
		{
			\begin{tabular}{l|c|c|c|c|c}
				\toprule


				Method&Estimator &RTE(cm) &RRE(\degree) &RR(\%) &Time(s) \\
				\hline\hline
				\noalign{\smallskip}

                3DFeat-Net \cite{yew20183dfeat}&RANSAC
                &25.9 &\bf0.25
                &96.0\%&3.4\\
          
				FCGF \cite{choy2019fully}  &RANSAC
				&9.5 & 0.30 
				&96.6\% &3.4 \\ 

                D3Feat \cite{bai2020d3feat} &RANSAC
				&7.2 & 0.30
				&\bf{99.8\%} &3.1\\ 
                
                Predator \cite{huang2021predator} &RANSAC
				&6.8 & 0.27
				&\bf{99.8\%} &5.2\\
                    
				CoFiNet \cite{yu2021cofinet}   &RANSAC
				&8.5  &0.41
				&\bf{99.8\%} &1.9\\

                SpinNet \cite{ao2021spinnet}   &RANSAC
				&9.9  &0.47
				&99.1\% &60.6\\

                
				GeoTrans. \cite{qin2023geotransformer}   &RANSAC
				&7.4  &0.27
				&\bf{99.8\%} &1.6\\

                MAC$^{\dagger}$ \cite{zhang20233d} &Graph
                &8.5  &0.40
				&99.5\% &-\\
                
                RIGA \cite{yu2024riga}   &RANSAC
				&13.5  &0.45
				&99.1\% &-\\

                 PointTruss$^{\dagger}$ \cite{wupointtruss} & Graph
                &\bf5.3  
                &0.43 
				&99.6\% &0.2\\

                
                TurboReg$^{\dagger}$ \cite{yan2025turboreg} & Graph
                &9.0  &0.47
				&98.6\% &-\\

                \cline{1-6}\noalign{\smallskip}
                RegFormer \cite{Liu_2023_ICCV} &-
                &8.4 &\bf0.24
                &\bf{99.8\%}&\bf0.1\\
            
                RegFormer++  &-
                &\bf6.6&0.26
                &\bf{99.8\%}&\bf0.1\\
              
                \bottomrule
			\end{tabular}
		}
	\end{center}
	
	\label{table:ransac}
	
\end{table}

\begin{figure}[t]
  \begin{center}
		\resizebox{1.0\columnwidth}{!}
		{
			\includegraphics[scale=1.00]{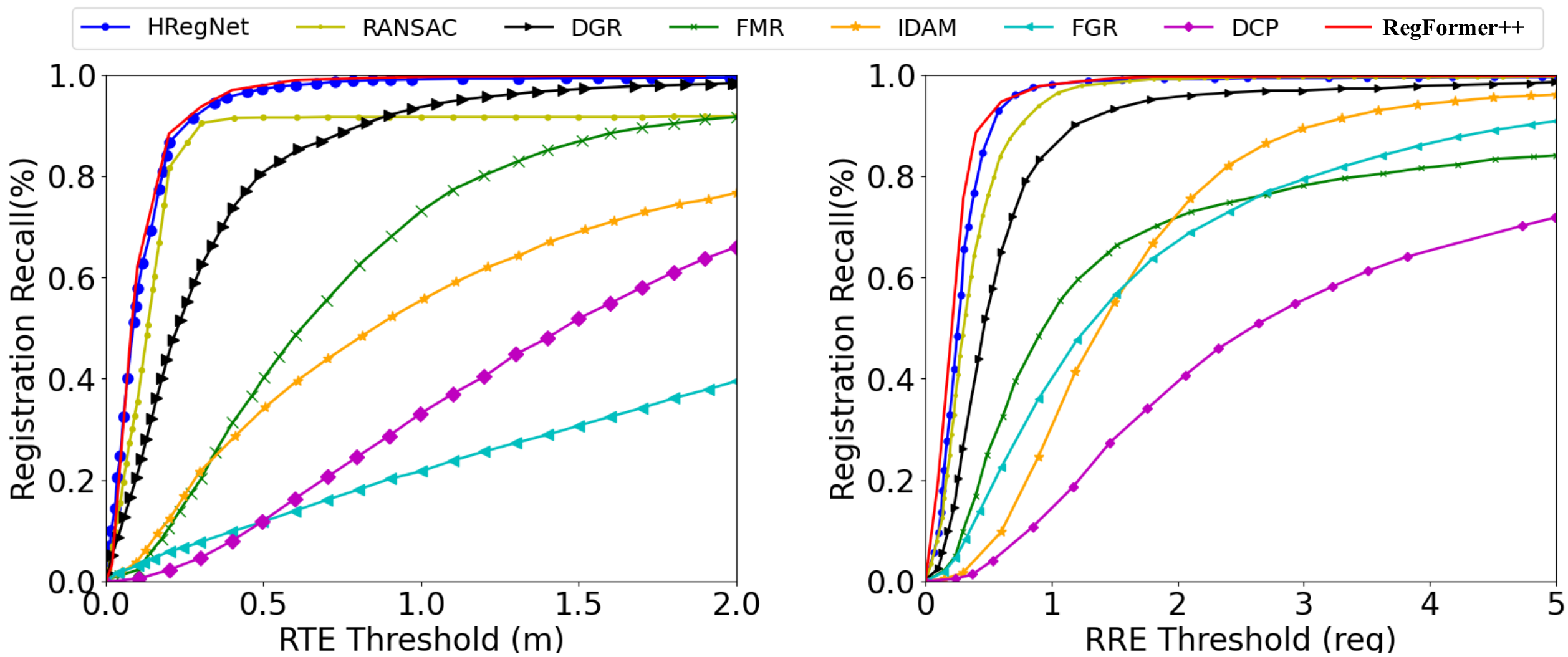}}
	\end{center}
    \vspace{-4mm}
  \caption{Registration recall with different RRE and RTE thresholds on the KITTI dataset.}
  \label{fig:rr}
\end{figure}

\setlength{\tabcolsep}{0.8mm}
\begin{table}[t]
	\centering
	\footnotesize
    \caption{Comparison results with state-of-the-arts on the NuScenes dataset. Input point pairs are 10-frame away prepared by \cite{lu2021hregnet}. The best performance is highlighted in bold. Registration Recall (RR) is defined as the success ratio where RRE$<$ 5\degree and RTE$<$ 2m. `NT' means normalized time per thousand points.}
    \vspace{-5mm}
	\begin{center}
		\resizebox{1.0\columnwidth}{!}
		{
			\begin{tabular}{l|c|c|c|c|c}
				\toprule

				
				Method &RTE(m) &RRE(\degree) &Recall(\%) &Time/Points&NT(ms) \\
				\hline\hline
				\noalign{\smallskip}

                FGR \cite{zhou2016fast}
                &0.71 
                &1.01 
                &32.2\%&284.6/11445&24.87\\

                RANSAC \cite{fischler1981random}
                &0.21 
                &0.74 
                &60.9\% &268.2/8192&32.74\\


                DCP \cite{wang2019deep}    
				&1.09  
				&2.07  
				&58.6\%  &45.5/1024&44.43 \\ 

                IDAM \cite{li2020iterative}    
				&0.47  
				&0.79 
				&88.0\% &32.6/4096&7.96 \\ 

                FMR \cite{huang2020feature}    
				&0.60  
				&1.61  
				&92.1\% &61.1/12000&5.09 \\

                DGR \cite{choy2020deep}    
				&0.21 & 0.48  
				&98.4\% &523.0/8192&63.84\\ 

                HRegNet \cite{lu2021hregnet}      
				&0.18 
				&0.45
				&99.9\%   &87.3/8192&10.66\\

                 LS-RegNet \cite{tao2025lsreg}
                 &0.15  &0.32
				&100\% &81.2/8192&9.91\\

                \cline{1-6}\noalign{\smallskip}    
                RegFormer \cite{Liu_2023_ICCV}
                &0.20
                &{0.22} 
                &{99.9\%} &\bf{85.6/60000}&\bf{1.42}\\

                RegFormer++
                &\bf0.09
                &\bf{0.18} 
                &\bf{100\%} &104.3/60000&1.74\\
                \bottomrule
			\end{tabular}
		}
	\end{center}
	\label{table:nuscenes}
	
\end{table}

\subsection{Quantitative Results on KITTI}
KITTI odometry dataset is composed of 11 sequences (00-10) with ground truth poses. Following the settings in \cite{choy2019fully,choy2020deep}, we use sequence 00-05 for training, sequence 06-07 for validation, and sequence 08-10 for testing. Also, the ground truth poses of KITTI are refined with ICP as in \cite{choy2019fully,bai2020d3feat,lu2021hregnet}.

\begin{figure*}
 \centering
 \includegraphics[width=1.0\linewidth]{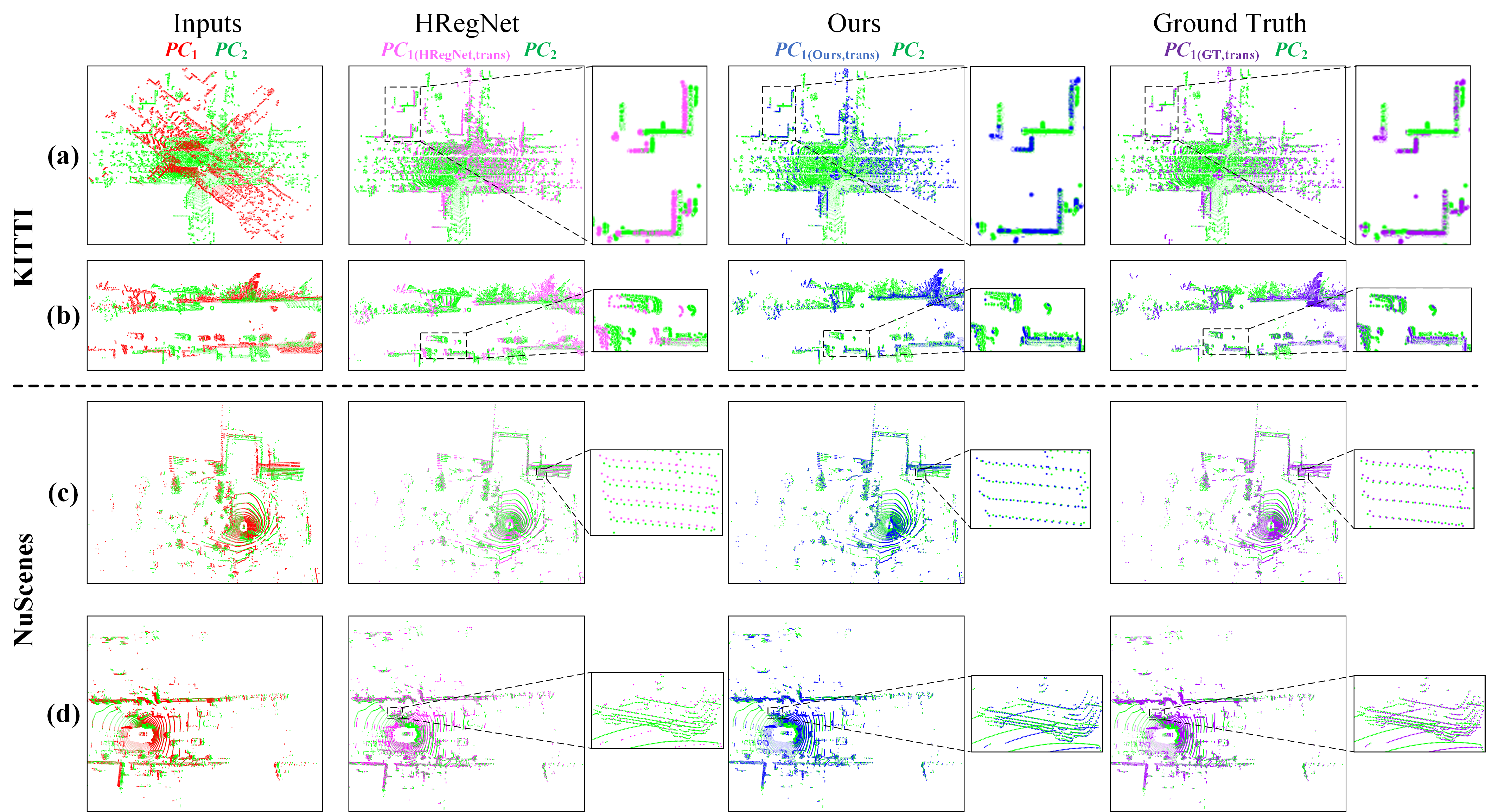}
 \vspace{-4mm}
 \caption{Registration results on KITTI and Nuscenes datasets. Point clouds colored red and green indicate the input \textcolor{red}{source} and \textcolor{green}{target} point clouds. Transformed source points by the \textcolor[rgb]{1,0,1}{estimated pose of HRegNet} and \textcolor{blue}{ours} are colored pink and blue respectively. The \textcolor[rgb]{0.667,0,1}{ground truth} is colored purple. Our RegFormer++ can align low-overlap input points even with large translations (b) and (d) or large rotations (a) and (c).}
 \label{fig:overlap}
\end{figure*}

\begin{figure*}
  \centering
  \includegraphics[width=1.00\linewidth]{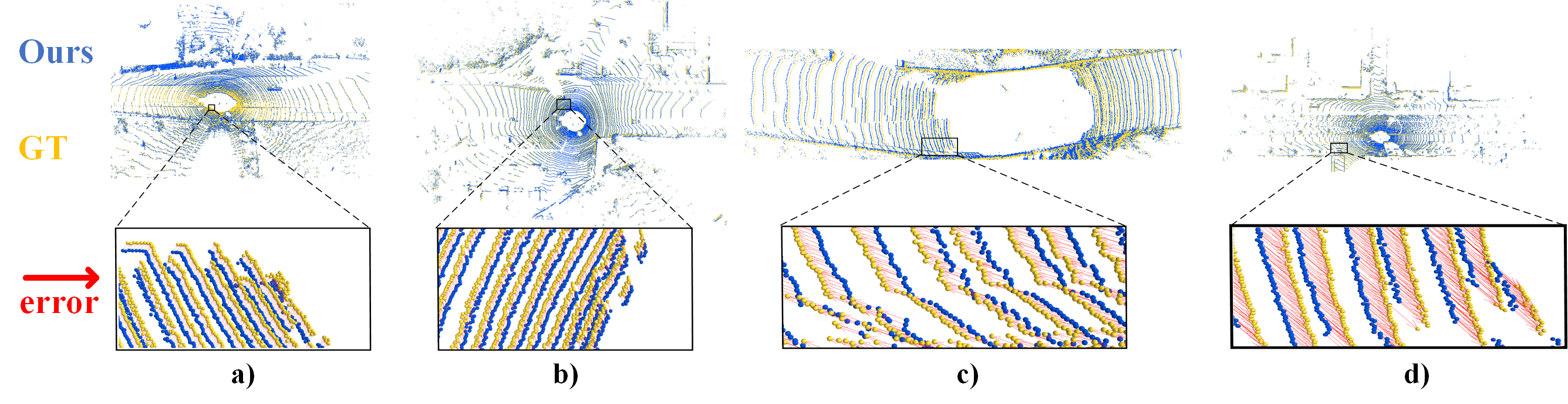}
  \vspace{-6mm}
  \caption{Visualization of registration errors. Point clouds colored yellow and blue indicate transformed source points by the ground truth (GT) and our estimated pose. Registration errors are visualized by a red vector pointing from estimated points to the GT ones.}
  
  \label{fig:error}
  
\end{figure*}


\vspace{3pt}\noindent\textbf{Comparison with State-of-the-Art Methods.} Following \cite{lu2021hregnet}, the current frame and the 10th frame after it are used to form input point pairs. We choose both traditional and learning-based registration methods for a comprehensive comparison. For classical methods, our network is superior to FGR \cite{zhou2016fast} by a large margin in both accuracy and efficiency. Compared to RANSAC \cite{fischler1981random}, our RegFormer++ has a 9.1\% higher RR. Also, RANSAC suffers from much lower efficiency (four times more total time than ours) due to the slow convergence. For learning-based methods, RegFormer++ is also compared with a series of state-of-the-art works. DCP \cite{wang2019deep}, IDAM \cite{li2020iterative}, FMR \cite{huang2020feature}, PPT \cite{wang2025point}, and LS-RegNet \cite{tao2025lsreg} are all feature-based registration networks that extract local descriptors. DGR \cite{choy2020deep} achieves competitive performance in indoor scenes with global features. As in Table~\ref{table:2m}, our RegFormer++ has much lower RRE and RTE, higher RR than all the above learning-based methods without designing discriminative descriptors. HRegNet \cite{lu2021hregnet} is a recent CNN-based SOTA method for efficient outdoor large-scale point cloud registration. Compared with HRegNet, our RegFormer++ is still more accurate in terms of all metrics (RTE, RRE, and RR). As for efficiency, both total time and normalized time are given. Normalized time is calculated by the processing speed per thousand points. As illustrated in Table~\ref{table:2m}, our method can process large-scale points with the highest average efficiency (0.82ms and 1.14ms respectively). Compared with the most efficient CNN-based method HRegNet, our model has a competitive total inference time and one-sixth of the average time. Registration recalls with different RRE and RTE thresholds are also displayed in Fig.~\ref{fig:rr}, which proves that our RegFormer++ is extremely robust to various threshold settings.

\setlength{\tabcolsep}{3mm}
\begin{table}[t]
	\centering
	\footnotesize
    \caption{Comparison results on the Argoverse dataset \cite{chang2019argoverse}. We follow the standard dataset settings to evaluate the performance and train both HRegNet and our method from scratch.}
    \vspace{-5mm}
	\begin{center}
		\resizebox{1.0\columnwidth}{!}
		{
    \begin{tabular}{l|cc|cc|c}
				\toprule
				&    \multicolumn{2}{c|}{RTE(m)} &\multicolumn{2}{c|}{RRE(\degree)} &   \\ 
				
				
				\multirow{-2}{*}{\begin{tabular}[c]{@{}c@{}}Method \end{tabular}}
				&  AVG  & STD   & AVG    & STD        &\multirow{-2}{*}{\begin{tabular}[c]{@{}c@{}}RR(\%) \end{tabular}}  \\
				\hline\hline
				\noalign{\smallskip}

                HRegNet \cite{lu2021hregnet}      
				&1.12 & 0.53
				&1.04 & 1.12
				&25.7\%   \\

                RegFormer++
                &\bf0.01 &\bf0.01
                &\bf{0.11} & \bf0.27
                &\bf{85.5\%} \\
                \bottomrule
			\end{tabular}
		}
	\end{center}	
	\label{table:argo}
\end{table}

\vspace{3pt}\noindent\textbf{Comparison with RANSAC-based or Graph-based Methods.} RANSAC is a commonly employed estimator for filtering outliers. With no need for RANSAC, our RegFormer++ leverages the attention mechanism for improving resilience to outliers by learning global features. Its effectiveness is demonstrated by the comparison with RANSAC-based methods in Table~\ref{table:ransac}. We follow settings in \cite{qin2022geometric} using input point pairs at least 10m away. Also, all methods are divided into two categories in terms of different backbones for cross-frame association: CNN and Transformer. Our network is on par with all SOTA CNN-based works including 3DFeatNet \cite{yew20183dfeat}, FCGF \cite{choy2019fully}, D3Feat \cite{bai2020d3feat}, CoFiNet \cite{yu2021cofinet}, Predator \cite{huang2021predator}, and SpinNet \cite{ao2021spinnet}. Compared with the recent SOTA method Predator, our RegFormer++ has lower RTE and RRE. As to efficiency, Predator has 52 $\times$ more runtime compared with ours. In terms of transformer-based networks, GeoTransformer \cite{qin2022geometric} introduces geometric features into transformer and has a marginally higher RRE, RTE, and also obvious efficiency decline. Our efficient RegFormer++ has a $16\times$  speed-up compared with theirs. RIGA \cite{yu2024riga} designs rotation-invariant and globally-aware descriptors where self and cross attention are utilized for global context aggregation. Our method has a significant improvement in terms of rotation and translation errors and also a successful registration ratio. We also compare our method with graph-based methods, where the maximal clique \cite{zhang20233d} algorithm is largely explored recently. Compared to MAC \cite{zhang20233d}, TurboReg \cite{yan2025turboreg}, and PointTruss \cite{wupointtruss}, our method achieves rather competitive accuracy.

\subsection{Quantitative Results on NuScenes}
We further evaluate our RegFormer++ on the NuScenes dataset. It consists of 1000 scenes, including 850 scenes for training and validation, and 150 scenes for testing. Following \cite{lu2021hregnet}, we use 700 scenes for training, 150 scenes for validation, and 10-frame away point pairs. 

As illustrated in Table~\ref{table:nuscenes}, our registration accuracy outperforms all classical works and learning-based ones. Our RegFormer++ has a 39.1\% RR improvement compared with RANSAC with only 38.9\% of their runtime. For learning-based methods, our model is superior to DCP \cite{wang2019deep}, IDAM \cite{li2020iterative}, and FMR \cite{huang2020feature} by a large margin. Compared with HRegNet \cite{lu2021hregnet}, our RegFormer++ has a 100\% registration recall, half lower RTE and  RRE than theirs. Moreover, their efficiency is also lower than ours, and their feature extraction module needs to be pre-trained.

\subsection{Quantitative Results on Argoverse}
We also evaluate the performance of our RegFormer++ on the Argoverse dataset \cite{chang2019argoverse}. The Argoverse benchmark collects data from six U.S. cities and incorporates a wide range of weather conditions and driving scenarios. Different from KITTI, its vertical viewing covers 40\degree  fields. As in Table \ref{table:argo}, our method outperforms HRegNet by a large margin. We keep the same dataset settings and train both models from scratch. We find that HRegNet cannot converge after a few epochs. In contrast, our method can still learn an accurate pose.

\setlength{\tabcolsep}{1mm}
\begin{table}[t]
	\centering
	\footnotesize
    \caption{Generalization results on the ETH dataset. All models are pre-trained on other datasets and then directly evaluated on the unseen ETH dataset \cite{pomerleau2012challenging}.}
    \vspace{-5mm}
	\begin{center}
		\resizebox{1.0\columnwidth}{!}
		{
     \begin{tabular}{l|cc|cc|c}
    \cline{1-6}
    & \multicolumn{2}{c|}{Gazebo} &\multicolumn{2}{c|}{Wood} & \\ 
				

				\multirow{-2}{*}{\begin{tabular}[c]{@{}c@{}} Method \end{tabular}} &Summer  &Winter   & Autumn    & Summer        &\multirow{-2}{*}{\begin{tabular}[c]{@{}c@{}}Avg. \end{tabular}}  \\
				\hline
				\noalign{\smallskip}   

                FPFH \cite{rusu2009fast}&38.6\% &14.2\%
                &14.8\% &20.8\%
                &22.1\%\\

                SHOT \cite{tombari2010unique}&73.9\% &45.7\%
                &60.9\% &64.0\%
                &61.1\%\\

                3DMatch \cite{zeng20173dmatch}&22.8\% &8.3\%
                &13.9\% &22.4\%
                &16.9\%\\

                CGF \cite{khoury2017learning}&37.5\% &13.8\%
                &10.4\% &19.2\%
                &20.2\%\\

                PerfectMatch \cite{gojcic2019perfect}&91.3\% &84.1\%
                &67.8\% &72.8\%
                &79.0\%\\

                FCGF \cite{choy2019fully}&22.8\% &10.0\%
                &14.8\% &16.8\%
                &16.1\%\\

                D3Feat \cite{bai2020d3feat}&85.9\% &63.0\%
                &49.6\% &48.0\%
                &61.6\%\\

                LMVD \cite{li2020end}&85.3\% &72.0\%
                &84.0\% &78.3\%
                &79.9\%\\

                SpinNet \cite{ao2021spinnet}&92.9\% &91.7\%
                &\bf92.2\% &94.4\%
                &92.8\%\\
                
                RegFormer++ &\bf93.2\% &\bf91.8\%
                &91.9\% &\bf94.9\%
                &\bf93.0\%\\
                
                \bottomrule
			\end{tabular}
		}
	\end{center}	
	\label{table:eth}
\end{table}

\setlength{\tabcolsep}{0.9mm}
\begin{table*}[t]
	\footnotesize
    \caption{Comparison with the state-of-the-art on the LiDAR odometry task. $t_{rel}$, $r_{rel}$ indicate the average translation RMSE (\%) and rotation RMSE ($^{\circ}$/100m) respectively on all subsequences in the length of $100,200,...,800m$. `$^*$' means that only odometry results without mapping are listed for a fair comparison. `$^{\dagger}$' means the evaluation sequences. `${NG}$' means results are not given. We re-evaluate the results of PWCLO \cite{wang2021pwclo} and obtain the same results as in \cite{ali2023delo}. The best results for each sequence are \textcolor{red}{\bf{bold}}, and the second best is \textcolor{blue}{\underline{underlined}}.}
    \vspace{-5mm}
	\begin{center}
		\resizebox{1.0\textwidth}{!}
		{
			\begin{tabular}{l|l|cc|cc|cc|cc|cc|cc|cc|cc|cc|cc|cc||cc}
				\toprule
				&&  \multicolumn{2}{c|}{00}  &\multicolumn{2}{c|}{01}      & \multicolumn{2}{c|}{02} & \multicolumn{2}{c|}{03} &  \multicolumn{2}{c|}{04} & \multicolumn{2}{c|}{05} & \multicolumn{2}{c|}{06} & \multicolumn{2}{c|}{07$^{\dagger}$} & \multicolumn{2}{c|}{08$^{\dagger}$} & \multicolumn{2}{c|}{09$^{\dagger}$} &\multicolumn{2}{c||}{10$^{\dagger}$} &\multicolumn{2}{c}{Mean on 07-10} \\ 
				\cline{3-26}
				
				\multirow{-2}{*}{\begin{tabular}[c]{@{}c@{}} \end{tabular}} & \multirow{-2}{*}{\begin{tabular}[c]{@{}c@{}}Method \end{tabular}}
				&  $t_{rel}$  & $r_{rel}$   & $t_{rel}$                       & $r_{rel}$               & $t_{rel}$                          & $r_{rel}$   & $t_{rel}$ & $r_{rel}$   & $t_{rel}$                          & $r_{rel}$   & $t_{rel}$ & $r_{rel}$    & $t_{rel}$                          & $r_{rel}$   & $t_{rel}$ & $r_{rel}$ & $t_{rel}$                          & $r_{rel}$   & $t_{rel}$ & $r_{rel}$    & $t_{rel}$ & $r_{rel}$   & $t_{rel}$ & $r_{rel}$      \\
				\hline\hline \noalign{\smallskip}

				&ICP-po2po    
				&6.88&	2.99
				&11.21&	2.58
				& 8.21	&3.39
				&11.07&	5.05
				& 6.64	&4.02
				& 3.97&	1.93
				&1.95	&1.59
				&5.17	&3.35
				&10.04	&4.93
				&6.93	&2.89
				&8.91	&4.74
				&7.36	&3.41
				\\ 
				
				&ICP-po2pl     
				&3.80 &	1.73
				&13.53 & 2.58 
				&9.00 & 2.74 
				&2.72 & 1.63 
				&2.96 & 2.58 
				&2.29 & 1.08 
				&1.77 & 1.00 
				&1.55 & 1.42 
				&4.42 & 2.14 
				&3.95 & 1.71 
				&6.13 & 2.60 
				&4.74 &1.93
                \\

                & GICP \cite{segal2009generalized}    
				&1.29 & 0.64 
				&4.39 & 0.91 
				&2.53 & 0.77 
				&1.68 & 1.08 
				&3.76 & 1.07 
				&1.02 & 0.54 
				&0.92 & 0.46 
				&0.64 &0.45 
				&1.58 & 0.75 
				&1.97 & 0.77 
				&1.31 &0.62
				&1.92 & 0.73
                \\

                &CLS \cite{velas2016collar}    
                &2.11 &	0.95 
                &4.22 &	1.05 
                &2.29 & 0.86 
                &1.63 &	1.09 
                &1.59 & 0.71 
                &1.98 &	0.92
                &0.92 & 0.46 
                &1.04 & 0.73 
                &2.14 & 1.05 
                &1.95 & 0.92
                &3.46 & 1.28 
                &2.15 &1.00
                \\

                &LOAM$^*$    \cite{zhang2017low} 
				&15.99  &6.25	  
				& 3.43 &0.93	 
				& 9.40 & 3.68 
				& 18.18 &9.91	  
				& 9.59 &  4.57
				& 9.16 &  4.10
				& 8.91 &  4.63
				& 10.87 &  6.76
				& 12.72 &  5.77 
				& 8.10 &  4.30
				& 12.67 &  8.79
				& 10.82 & 5.43
				\\         
                
                
                &LeGO-LOAM \cite{shan2018lego}    
                &2.17 &	1.05 
                &13.4 &	1.02 
                &2.17 & 1.01 
                &2.34 &	1.18 
                &1.27 & 1.01 
                &1.28 &	0.74 
                &1.06 & 0.63 
                &1.12 & 0.81 
                &1.99 & 0.94 
                &1.97 & 0.98 
                &2.21 & 0.92 
                &2.49 & 1.00
                \\
                
                &SuMa$^*$ \cite{behley2018efficient}    
                &2.93 &	0.92
                &2.09 &	0.93 
                &4.05 & 1.22 
                &2.30 &	0.79 
                &1.43 & 0.75
                &11.9 &	1.06
                &1.46 & 0.79 
                &1.75 & 1.17 
                &2.53 & 0.96 
                &1.92 & 0.78 
                &1.81 & 0.97 
                &2.93 & 0.92
                \\
                
                \multirow{-8}{*}{\begin{tabular}[c]{@{}c@{}}\rotatebox{90}{Classic} \end{tabular}}&PUMA \cite{vizzo2021poisson}    
                &1.46 &	0.68 
                &3.38 &	1.00
                &1.86 & 0.72 
                &1.60 &	1.10
                &1.63 & 0.92 
                &1.20 &	0.61 
                &0.88 & 0.42 
                &0.72 & 0.55 
                &1.44 & \textcolor{blue}{\underline{0.61}} 
                &1.51 & 0.66
                &1.38 & 0.84
                &1.55 & 0.74
                \\
                \hline \noalign{\smallskip}

                &Zhou et al. \cite{zhou2017unsupervised}    
				&NG &NG
				&NG &	NG
				&NG & NG
				&NG &	NG
				&NG & NG
				&NG &	NG
				&NG & NG
				&21.3 & 6.65
				&21.9 &2.91
				&18.8 &3.21
				&14.3 & 3.30
				&19.10 & 4.02
				\\ 

                &LO-Net \cite{li2019net}    
				&1.47 & 0.72 
				&1.36 & 0.47  
				&1.52 & 0.71 
				&\textcolor{blue}{\underline{1.03}} &\textcolor{blue}{\underline{0.66}} 
				&\textcolor{blue}{\underline{0.51}} & 0.65
				&1.04 & 0.69 
				&0.71 & 0.50 
				&1.70 & 0.89 
				&2.12 & 0.77 
				&1.37 &0.58 
				&1.80 & 0.93 
				&1.33 & 0.69
				\\ 
			
                &SelfVoxeLO \cite{2010.09343}    
				&NG & NG 
				&NG & NG
				&NG & NG 
				&NG & NG 
				&NG & NG 
				&NG & NG 
				&NG & NG 
				&2.51 &1.15 
				&2.65 &1.00 
				&2.86 &1.17 
				&3.22 &1.26
				&2.81 &1.15
				\\ 
                
                &RSLO$^*$ \cite{xu2022robust}    
                &NG & NG 
				&NG & NG
				&NG & NG 
				&NG & NG 
				&NG & NG 
				&NG & NG 
				&NG & NG 
				&2.37 &1.15 
				&2.14 &0.92 
				&2.61 &1.05
				&2.33 &0.94
				&2.36 &1.02
                \\
                &PWCLO \cite{wang2021pwclo} 
                &0.89 &0.43 
                & \textcolor{red}{\bf{1.11}} &\textcolor{blue}{\underline{0.42}} 
				&1.87 &0.76 
				&1.42 &  0.92 
				&1.15 & 0.94 
				&1.34 & 0.71 
				&\textcolor{blue}{\underline{0.60}} & 0.38 
				&1.16 & 1.00 
				&1.68 &0.72 
				&\textcolor{red}{\bf{0.88}} &\textcolor{blue}{\underline{0.46 }}
				&2.14 &0.71
				&1.47 &0.72

                 \\
                &DELO \cite{ali2023delo}   
                &1.43 &0.81 
                &2.19 &0.57 
				&1.48 &\textcolor{blue}{\underline{0.52}} 
				&1.38 & 1.10 
				&2.45 & 1.70 
				&1.27 & 0.64 
				&0.83 & 0.35 
				&0.58 & \textcolor{red}{\bf{0.41}}  
				&1.36 &0.64 
				&1.23 &0.57 
				&1.53 &0.90
				&1.18 &0.63
                 \\
                &TransLO \cite{liu2023translo}   
                &\textcolor{red}{\bf{0.85} }&\textcolor{red}{\bf{0.38}}
                & \textcolor{blue}{\underline{1.16}} &0.45
				&\textcolor{red}{\bf{0.88}} &\textcolor{blue}{\underline{0.34}} 
				&\textcolor{red}{\bf{1.00}} &  0.71 
				&\textcolor{red}{\bf{0.34}} & \textcolor{red}{\bf{0.18}} 
				&\textcolor{blue}{\underline{0.63}} & \textcolor{blue}{\underline{0.41}}
				&0.73 & \textcolor{blue}{\underline{0.31}}
				&\textcolor{blue}{\underline{0.55}} & \textcolor{blue}{\underline{0.43}} 
				&\textcolor{blue}{\underline{1.29}} &\textcolor{red}{\bf{0.50}} 
				&\textcolor{blue}{\underline{0.95}} &\textcolor{blue}{\underline{0.46}}
				&\textcolor{blue}{\underline{1.18}}
                &\textcolor{blue}{\underline{0.61}}
				&\textcolor{blue}{\underline{0.99}} &\textcolor{blue}{\underline{0.50}}
				
                \\
                	      					
				\multirow{-8}{*}{\begin{tabular}[c]{@{}c@{}}\rotatebox{90}{Learning} \end{tabular}}&RegFormer++      
				&\textcolor{blue}{\underline{0.86}} &\textcolor{blue}{\underline{0.39}}
				&1.34 & \textcolor{red}{\bf{0.40}}
				&\textcolor{blue}{\underline{1.10}} & \textcolor{red}{\bf{0.33}}
				&\textcolor{red}{\bf1.00} &\textcolor{red}{\bf{0.55}}
				&0.66 & \textcolor{blue}{\underline{0.37}}
				&\textcolor{red}{\bf0.61} & \textcolor{red}{\bf{0.36}}
				&\textcolor{red}{\bf0.46} & \textcolor{red}{\bf0.29}
				&\textcolor{red}{\bf0.53} & \textcolor{blue}{\underline{0.43}}
				&\textcolor{red}{\bf{1.25}} & 0.63
				&1.02 & \textcolor{red}{\bf{0.32}}
				&\textcolor{red}{\bf{1.04}} & \textcolor{red}{\bf{0.53}}
				&\textcolor{red}{\bf0.96}  &\textcolor{red}{\bf{0.48}}

				\\  \bottomrule
			\end{tabular}
		}
	\end{center}
	
	\label{table:lidar}
\end{table*}

\begin{figure*}
  \centering
  \includegraphics[width=1.00\linewidth]{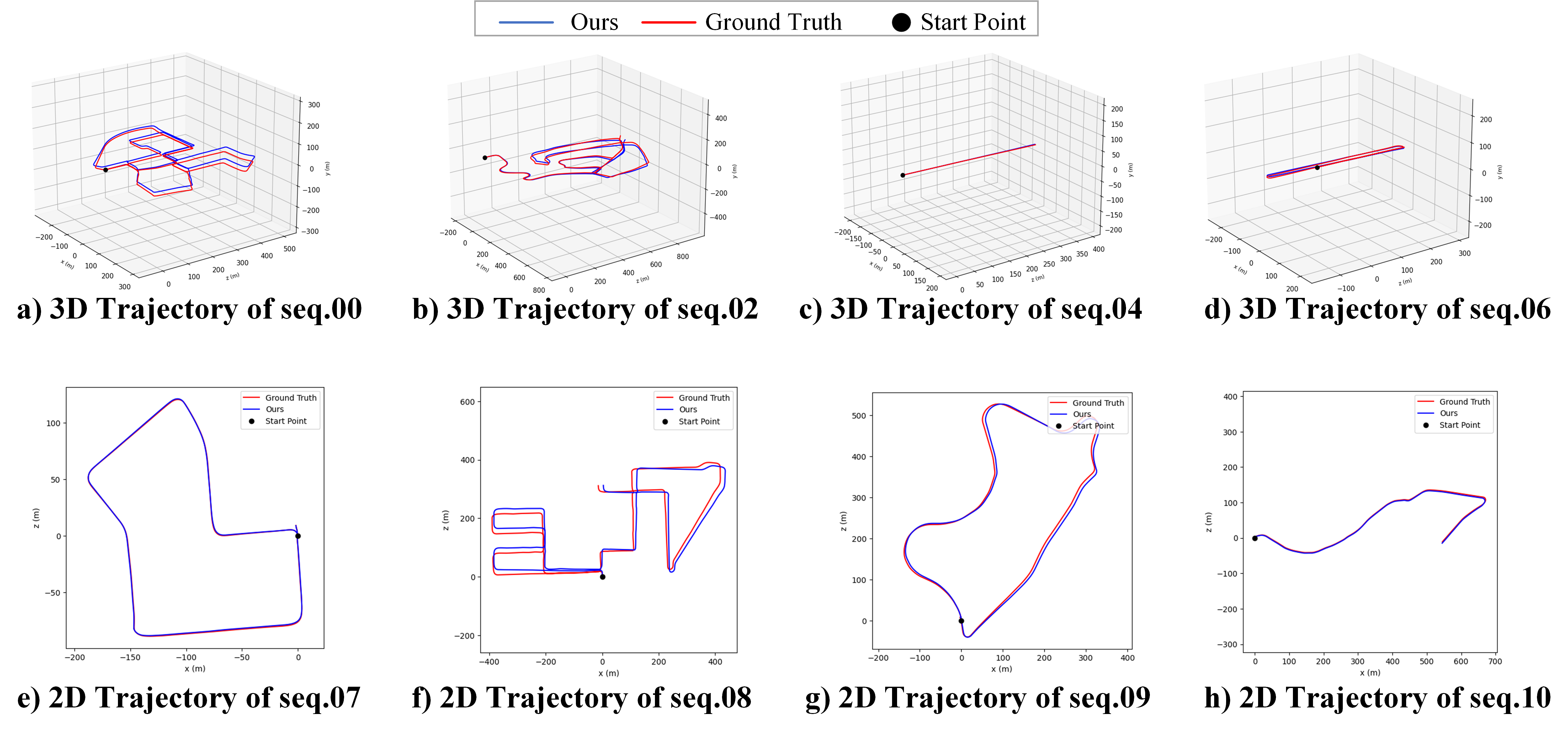}
  \vspace{-6mm}
  \caption{Visualization of 2D and 3D trajectories of LiDAR odometry. The blue and red lines are respectively our estimated and the ground truth trajectory.}
  \label{fig:trajectory}
\end{figure*}
\subsection{Qualitative Visualization}
\textbf{Low-Overlap Registration.} Fig.~\ref{fig:overlap} selects four challenging samples of registration results on the KITTI and NuScenes datasets. Our RegFormer++ can effectively align source and target point clouds even though they originally have large translations or rotations with low overlap. Also, our RegFormer++ has higher registration accuracy compared with HRegNet \cite{lu2021hregnet}, where transformed $PC_1$ of our method is almost overlapped with $PC_2$. 

\vspace{3pt}\noindent\textbf{Visualization of Registration Errors.} To further study the source of registration errors and how they are distributed, error vectors are visualized in Fig.~\ref{fig:error}. We find that our registration errors are also influenced by the surroundings due to the data-driven characteristics. When features on both sides of the vehicle are sufficient as Fig.~\ref{fig:error} a) and b), errors are relatively smaller and distributed evenly. It can be attributed to the structured buildings around, which offer solid positioning information. However, if there are scarce reference objects besides the car or surrounding features are monotonous as in Fig.~\ref{fig:error} c) and d), larger errors can present, mainly because features along the road are relatively monotonous, and the vehicle cannot accurately identify its forward movement without sufficient supporting information.


\subsection{Generalization on the Unseen Dataset}
We also evaluate the generalization capability of our RegFormer++ on the unseen ETH dataset \cite{pomerleau2012challenging}, which consists of four scenes, i.e., Gazebo-Summer, Gazebo-Winter, Wood-Summer, and Wood-Autumn. Different from KITTI, Argoverse, and NuScenes datasets which are based on mechanical rotating liDAR Velodyne, ETH is captured by static terrestrial scanners. Even so, our method can also generalize well on this type of unseen data. We directly test the registration recall with the pre-trained model on the KITTI dataset. We compare our method with both hand-crafted descriptors FPFH \cite{rusu2009fast}, SHOT \cite{tombari2010unique}, and also learnable descriptors like 3DMatch \cite{zeng20173dmatch}, CGF \cite{khoury2017learning}, PerfectMatch \cite{gojcic2019perfect}, FCGF \cite{choy2019fully}, D3Feat \cite{bai2020d3feat}, LMVD \cite{li2020end}, and SpinNet \cite{ao2021spinnet} in Table~\ref{table:eth}. All these indoor registration methods are pre-trained on the 3DMatch dataset and then directly evaluated on ETH.  Recent method SpinNet \cite{ao2021spinnet} designs a rotation-invariant local descriptor that has a strong generalization ability for indoor point cloud registration. As in Table~\ref{table:eth}, compared to SpinNet, our method achieves an even better averaged performance with a 0.2\% improvement. This proves that our method can also adapt to different sensors. 
\begin{figure}
  \centering
  \includegraphics[width=1.00\linewidth]{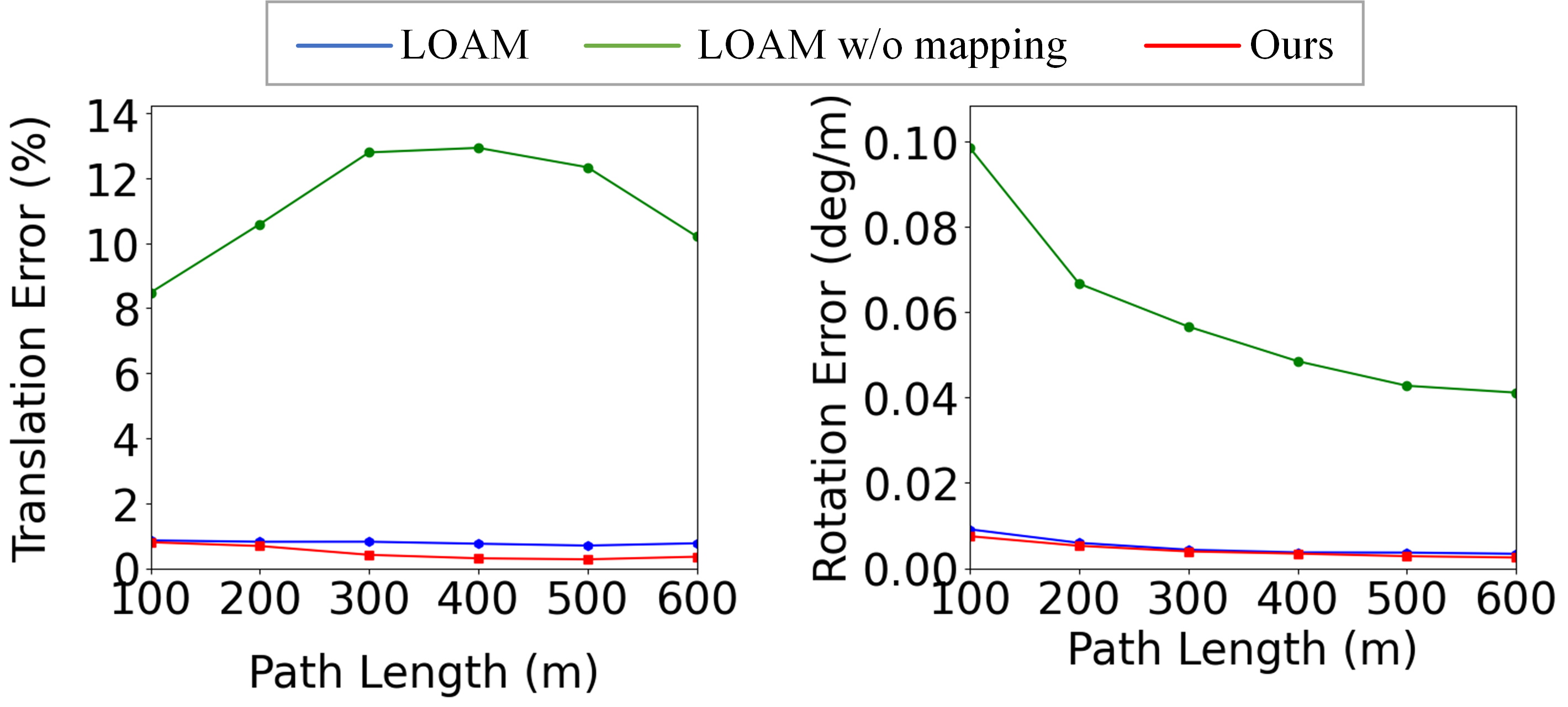}
  \vspace{-6mm}
  \caption{Comparison results with traditional LiDAR odometry LOAM. }
  \label{fig:loam}
\end{figure}

\subsection{Generalization on LiDAR Odometry Task}
For evaluating the generalization ability on real-world tasks, we also apply our large-scale registration network RegFormer++ to the real-time LiDAR odometry, which belongs to the font-end of SLAM system. From Table~\ref{table:lidar}, our RegFormer++ can outperform all recent learning-based methods, including LO-Net \cite{li2019net}, PWCLO \cite{wang2021pwclo}, TransLO \cite{liu2023translo}, DELO \cite{ali2023delo} et al, by a large margin. Compared with the recent classical method PUMA \cite{vizzo2021poisson}, our network has a 38.1\% lower translation error and a 35.1\% lower rotation error on the evaluation sequences 07-10. With respect to learning-based methods, our RegFormer++ also outperforms the recent transformer-based SOTA method TransLO \cite{liu2023translo}. Notably, loop closure and mapping are not included in our registration network, so we fairly compare with LOAM \cite{zhang2017low} and SUMA \cite{behley2018efficient} without mapping and loop closure. We visualize the 2D and 3D trajectories of a series of sequences in Fig.~\ref{fig:trajectory}, where our estimated pose can accurately track the ground truth one. Both quantitative and qualitative results show that our RegFormer++ has a strong generalization ability for downstream applications. In Fig. \ref{fig:loam}, we also report estimated trajectories and errors compared to LOAM. Our designed algorithm not only outperforms the odometry-only LOAM by a large margin, but also surpasses LOAM with mapping.


\subsection{Ablation Study}
In this section, extensive ablation studies are conducted for the effectiveness evaluation of each designed element.

\vspace{3pt}\noindent\textbf{Hierarchical Architectures.} We adopt a coarse-to-fine architecture for the precise pose estimation. To validate the effectiveness of the hierarchical design, we separately output the estimated transformation matrices from coarser layers and re-evaluate the metrics. As displayed in Table~\ref{table:ablation1}, rotation and translation are estimated from layer 3 (a), layer 2 (b), and layer 1 (c). It is obvious that registration errors get smaller as the transformation is iteratively refined. 

\setlength{\tabcolsep}{1.3mm}
\begin{table}[t]
	\centering
	\footnotesize
    \caption{Ablation studies of the hierarchical architecture. }
    \vspace{-5mm}
	\begin{center}
		\resizebox{1.0\columnwidth}{!}
		{
			\begin{tabular}{l|c|c|c}
				\toprule


				Model &RTE(m) &RRE(\degree) &Recall(\%) \\
				\hline\hline
				\noalign{\smallskip}

                (a) Transformation $q_3$, $t_3$
                &0.79 $\pm$ 0.47& 1.46 $\pm$ 0.98
                &74.9\%\\
          
				(b) Transformation $q_2$, $t_2$
				&0.53 $\pm$ 0.37 &1.37 $\pm$0.91
				&88.9\%  \\ 

                (c) Transformation $q_1$, $t_1$ 
				 &0.20 $\pm$ 0.19& 0.57 $\pm$ 0.50
                &99.7\%\\

                (d) Transformation $q_0$, $t_0$
                &\bf{0.04 $\pm$ 0.06} &\bf{0.14 $\pm$ 0.13}
                &\bf{100\%}\\

                \bottomrule
			\end{tabular}
		}
	\end{center}
   
		
	\label{table:ablation1}
	
\end{table}
\setlength{\tabcolsep}{0.8mm}
\begin{table}[t]
	\centering
	\footnotesize
    \caption{Ablation studies of components in Point Swin Transformer (PST). }
    \vspace{-5mm}
	\begin{center}
		\resizebox{1.0\columnwidth}{!}
		{
			\begin{tabular}{l|c|c|c}
				\toprule


				Model &RTE(m) &RRE(\degree) &Recall(\%) \\
				\hline\hline
				\noalign{\smallskip}

          



                (a) w/o projection mask
                &0.14 $\pm$ 0.13&0.41 $\pm$ 0.43
                &99.5\%\\


                (b) replaced by 2D CNN
                &0.51 $\pm$ 0.39&0.76 $\pm$ 0.73
                &83.6\%\\

                (c) replaced by PointNet++ \cite{qi2017pointnet++}
                &0.21 $\pm$ 0.20&0.47 $\pm$ 0.51
                &93.6\%\\
                (d) replaced by Point transformer \cite{zhao2021point}
                &0.11 $\pm$ 0.14&0.24 $\pm$ 0.26
                &99.6\%\\

                \cline{1-4}\noalign{\smallskip}

                Ours (Full)
                &\bf{0.04 $\pm$ 0.06} &\bf{0.14 $\pm$ 0.13}
                &\bf{100\%}\\

                \bottomrule
			\end{tabular}
		}
	\end{center}

	\label{table:ablation2}
	
\end{table}

\vspace{3pt}\noindent\textbf{Projection Masks.} The projection mask in our transformer enables our network to be aware of invalid projected 2D positions. As in Table~\ref{table:ablation2} (a), the registration accuracy drops dramatically when this mask is removed, since numerous invalid pixels are also taken into account.

\vspace{3pt}\noindent\textbf{Point Swin Transformer.} Different from most previous works, our RegFormer++ focuses on more global features with transformer. The global modeling ability enables our network to sufficiently capture dynamics and recover the location of occluded objects. To verify this, we replace the feature extraction transformer with CNNs, keeping other components unchanged. As indicated in Table~\ref{table:ablation2} (b), 2D CNN is used to extract local features with the same downsampling scale. PointNet++ \cite{qi2017pointnet++} is also utilized to replace our transformer in Table~\ref{table:ablation2} (c). The results show both local features extracted by 2D CNN and Pointnet++ have larger registration errors since their receptive fields are constrained. In the paper, we embed geometric features into our 2D transformer through a kernel-based feature embedding layer. Here, we also replace our feature extraction module with an existing 3D transformer \cite{zhao2021point}, which witnesses a 0.4\% recall decline.

\setlength{\tabcolsep}{0.6mm}
\begin{table}[t]
	\centering
	\footnotesize
    \caption{Ablation studies of components in Bijective Association Transformer (BAT) module. }
    \vspace{-5mm}
	\begin{center}
		\resizebox{1.0\columnwidth}{!}
		{
			\begin{tabular}{l|c|c|c}
				\toprule


				Model &RTE(m) &RRE(\degree) &Recall(\%) \\
				\hline\hline
				\noalign{\smallskip}

                (a) w/o cross attention
                &0.09$\pm$ 0.11&0.20 $\pm$ 0.29
                &98.8\%\\

                (b) w/o all-to-all point gathering
                &0.86$\pm$ 0.39&1.69 $\pm$ 1.37
                &67.2\%\\

                (c) replaced with cost volume in \cite{wang2021pwclo}
                &0.69 $\pm$ 0.33&0.98 $\pm$ 0.82 &71.2\%\\

               (d) replaced with cost volume in \cite{wang2022matters}
                &0.15 $\pm$ 0.17 &0.21 $\pm$ 0.25
                &94.8\%\\

                \cline{1-4}\noalign{\smallskip}

                Ours (Full)
                &\bf{0.04 $\pm$ 0.06} &\bf{0.14 $\pm$ 0.13}
                &\bf{100\%}\\

                \bottomrule
			\end{tabular}
		}
	\end{center}
    
		
	\label{table:ablation3}
	
\end{table}


               



	
\setlength{\tabcolsep}{3mm}
\begin{table}[t]
	\centering
	\footnotesize
    \caption{Ablation studies of the transformed features in the optimal transport (OT) layer. }
    \vspace{-5mm}
	\begin{center}
		\resizebox{1.0\columnwidth}{!}
		{
			\begin{tabular}{c|c|c|c|c}
				\toprule


				$SF$&FE$_{trans}$ &RTE(m) &RRE(\degree) &Recall(\%) \\
				\hline\hline
				\noalign{\smallskip}

                &&0.07 $\pm$ 0.12& 0.20 $\pm$ 0.21
                &99.7\%\\
          
				\checkmark &
				&0.06 $\pm$ 0.10 &0.19 $\pm$0.17
				&99.8\% \\ 
    
                & \checkmark
                &0.05 $\pm$ 0.06 &0.18 $\pm$ 0.15
                &99.9\%\\

                \checkmark &\checkmark 
                &\bf{0.04 $\pm$ 0.06} &\bf{0.14 $\pm$ 0.13}
                &\bf{100\%}\\

                \bottomrule
			\end{tabular}
		}
	\end{center}
	\label{table:ablation5}
\end{table}


\vspace{3pt}\noindent\textbf{Bijective Association Transformer.} In this paper, cross-attention is leveraged to exchange contextual information between frames in advance. Here, we remove the cross-attention in BAT to quantitatively test the effectiveness. Table~\ref{table:ablation3} (a) shows that registration errors double without cross-attention module. All-to-all association layer on the coarsest layer is also extremely crucial to finding reliable correspondences and reducing mismatches as in Table~\ref{table:ablation3} (b).  Furthermore, we conduct experiments by replacing BAT with the cost volume mechanism \cite{wang2021pwclo, wang2022matters}, which is commonly used for consecutive frame association. From the results in Table~\ref{table:ablation3} (c) (d), we can witness at least 0.11m larger RTE and 5.2\% RR drop. 

\vspace{3pt}\noindent\textbf{Feature-Transformed Optimal Transport.} We also design a feature-level optimal transport module. Without the optimal transport, there are almost double registration errors and 0.3\% recall decline as in Table~\ref{table:ablation5}. Additionally, there is an obvious decline of the deviations of both RTE and RRE metrics, demonstrating effectiveness of the deterministic optimal transport module in ensuring feature reliability. We also mix both transformed motion embeddings $FE_{trans}$ and transformed flow $SF$ together into initial motion embeddings $FE$, to obtain the final motion embeddings $ME$. In Table~\ref{table:ablation5}, we compare registration accuracy respectively without these two modules.

\begin{figure}[t]
  \centering
  \includegraphics[width=1.00\linewidth]{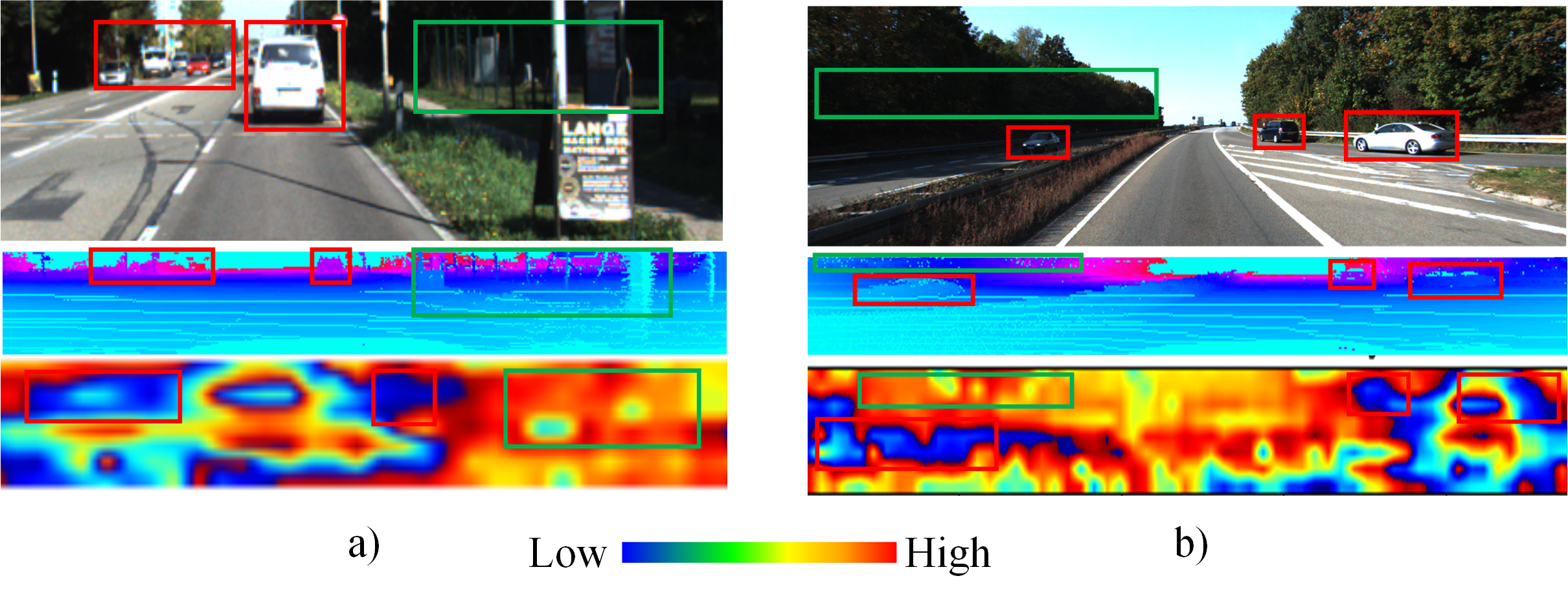}
  \vspace{-6mm}
  \caption{Visualization of attention weights. We give two samples here, where the first two rows respectively represent corresponding pictures and projected point clouds. Attention weights are visualized in the last row. Dynamic objects (red box) have lower attention weights, and static objects (green box) have higher weights.}
  \label{fig:weight}
  
\end{figure}

\section{Discussion}

In this section, we analyze and discuss why our RegFormer++ has such excellent performance even without subtly designed descriptors and a specialized outlier removal strategy. The competitive accuracy can be basically attributed to our outlier elimination capability and mismatch rejection strategy. We will elaborate on these two mechanisms in detail, respectively.

\vspace{3pt}\noindent\textbf{Global Modeling Ability to Filter Outliers.} Transformer can learn patch similarity globally while dynamics and occlusion have inconsistent global motion. Therefore, our transformer-based pipeline can effectively recognize and eliminate interference from these objects by paying less attention to these patches as in Fig.~\ref{fig:weight}. From the figure, dynamic cars tend to have lower attention weights contributing to the pose regression. In contrast, the surrounding static trees and buildings are assigned higher attention scores, which facilitate the final pose estimation. In this case, our RegFormer++ can maintain high registration accuracy even without explicit outlier rejections like RANSAC.

\begin{figure}[t]
  \centering
  \includegraphics[width=1.00\linewidth]{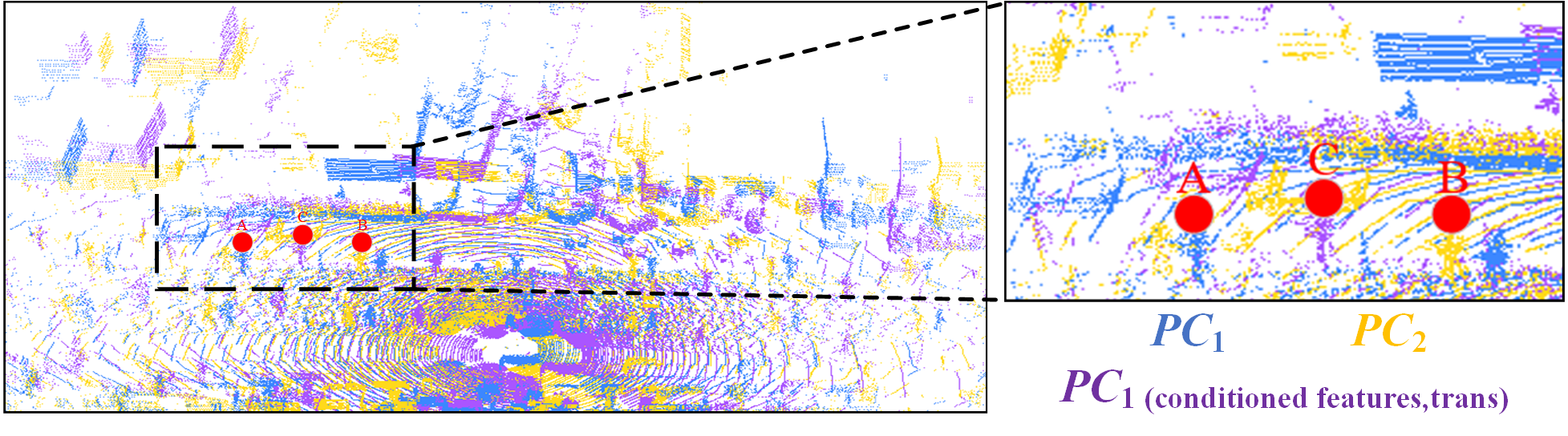}
  \vspace{-6mm}
  \caption{Visualization of the cross attention mechanism in BAT. Points A and B are corresponding points respectively in the source and target frame. Point C is the transformed position of A by conditioned features after cross-attention block.}
  \vspace{-3mm}
  \label{fig:visual2}
  
\end{figure}
 

\vspace{3pt}\noindent\textbf{Cross-Attention Mechanism for Reducing Mismatches.} In our Bijective Association Transformer module, a cross-attention block is first applied to exchange contextual information and interact motion embeddings between two frames. Here, we remove the rest parts of BAT, leveraging only conditioned features from the cross-attention block to generate a generally correct but not precise transformation. Then, it is used to transform input point clouds as in Fig.~\ref{fig:visual2} (purple). For each point in the source point cloud (blue), its corresponding point in the target one (yellow) is originally almost 10m away. Cross-attention can effectively shorten this distance between two frames by learning preliminary contextual features.

\section{Conclusion}
In this paper, we proposed a transformer-based large-scale registration network termed RegFormer++. Global features are extracted by 2D transformer with larger receptive fields to filter outliers. To cope with the irregularity and sparsity of raw point clouds, we leverage cylindrical projection to organize them orderly and present a projection mask to remove invalid pixels. Furthermore, a bijective association transformer, including a cross-attention-based contextual information exchange and the all-to-all point gathering, is designed to reduce mismatches. The whole model is RANSAC-free, keypoint-free, correspondence-free, and extremely accurate and efficient. We also demonstrate its great generalization ability on other LiDAR sensors and downstream tasks.

\small{\textbf{Data Availability Statement:}} All datasets are publicly available. Code is available at \url{https://github.com/IRMVLab/RegFormer}. 
%
%

\bibliographystyle{spmpsci}      

\bibliography{
bib/intro
}  


\end{document}